\documentclass{article}

\usepackage[preprint]{Styles/neurips-2025}

\usepackage{amsmath}
\usepackage{graphicx}
\usepackage{multirow}
\usepackage{enumitem}
\usepackage{wrapfig}
\usepackage[flushleft]{threeparttable}
\usepackage{makecell}

\usepackage[utf8]{inputenc} 
\usepackage[T1]{fontenc}    
\usepackage{hyperref}       
\usepackage{url}            
\usepackage{booktabs}       
\usepackage{amsfonts}       
\usepackage{nicefrac}       
\usepackage{microtype}      
\usepackage{xcolor}         
\usepackage{graphicx}  
\usepackage{subfig}    

\usepackage{wrapfig}

\usepackage{xspace}
\usepackage{array}

\usepackage{diagbox}

\usepackage{wrapfig}

\usepackage{algorithmicx,algorithm}

\usepackage{graphicx}
\usepackage{subcaption}

\PassOptionsToPackage{numbers, compress}{natbib}

\title{DBLoss: Decomposition-based Loss Function\\ for Time Series Forecasting}

\usepackage{amsmath,amsfonts,bm}

\def\eqref#1{equation~\ref{#1}}

\def\1{\bm{1}}

\def\mX{{\bm{X}}}
\def\mY{{\bm{Y}}}

\DeclareMathAlphabet{\mathsfit}{\encodingdefault}{\sfdefault}{m}{sl}
\SetMathAlphabet{\mathsfit}{bold}{\encodingdefault}{\sfdefault}{bx}{n}

\author{%
  Xiangfei Qiu$^{1}$, Xingjian Wu$^{1}$, Hanyin Cheng$^{1}$, Xvyuan Liu$^{1}$\\ \textbf{ Chenjuan Guo$^{1}$, Jilin Hu$^{1,2}$\thanks{Corresponding Author}, Bin Yang$^{1}$}\\
  $^1$East China Normal University, $^2$KLATASDS-MOE\\
\texttt{\{xfqiu, xjwu, hycheng, xyliu\}@stu.ecnu.edu.cn}, \\
\texttt{\{cjguo, jlhu, byang\}@dase.ecnu.edu.cn}
}

\begin{document}

\maketitle

\begin{abstract}
Time series forecasting holds significant value in various domains such as economics, traffic, energy, and AIOps, as accurate predictions facilitate informed decision-making. However, the existing Mean Squared Error (MSE) loss function sometimes fails to accurately capture the seasonality or trend within the forecasting horizon, even when decomposition modules are used in the forward propagation to model the trend and seasonality separately. To address these challenges, we propose a simple yet effective \textbf{\underline{D}}ecomposition-\textbf{\underline{B}}ased \textbf{Loss} function called \textbf{DBLoss}. This method uses exponential moving averages to decompose the time series into seasonal and trend components within the forecasting horizon, and then calculates the loss for each of these components separately, followed by weighting them. As a general loss function, DBLoss can be combined with any deep learning forecasting model. Extensive experiments demonstrate that DBLoss significantly improves the performance of state-of-the-art models across diverse real-world datasets and provides a new perspective on the design of time series loss functions.
\end{abstract}
\begin{center}
\textbf{Resources:} \href{https://github.com/decisionintelligence/DBLoss}{https://github.com/decisionintelligence/DBLoss}.
\end{center}

\def\HJL#1{\textcolor{green}{[Jilin: #1]}}

\section{Introduction}

Time Series Forecasting holds significant value in various domains such as economics~\citep{wang2025timeo1,li2025towards,ma2025robust,liu2025efficient}, traffic~\citep{freeformer,ma2025less,ma2025causal,liu2025timecma}, energy~\citep{wang2025optimal,huang2025mafs,ma2025mobimixer,miao2024less}, and AIOps~\citep{wang2025mitigating,yue2024sub,ma2025bist,wu2024catch}, as accurate predictions facilitate astute decision-making. To pursue accurate predictions, recent progress in Long-term Time Series Forecasting focuses on effectively capturing the inherent seasonality and trend, which reflect the changing laws of the time series, i.e., the inductive bias. Recently, dozens of deep learning models have been designed from light-weight to multi-scale, such as DLinear~\citep{DLinear}, OLinear~\citep{yue2025olinear}, CycleNet~\citep{lincyclenet}, TimesNet~\citep{TimesNet}, TimeBase~\citep{huang2025timebase}, PDF~\citep{PDFliu}, TimeMixer~\citep{TimeMixer}, and DUET~\citep{DUET}, which are aiming at capturing such inductive bias consistently within data for more accurate predictions.

\begin{figure*}[!tbp]
    \centering
    \includegraphics[width=1\linewidth]{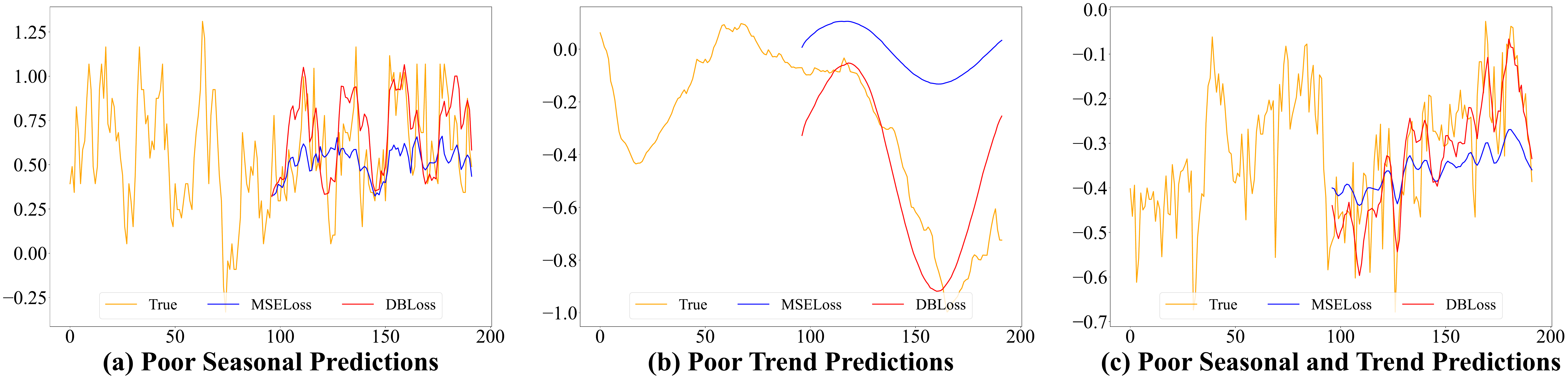}
    \caption{Limitations of MSE loss in capturing seasonality or trend within the forecasting horizon.}
\label{fig: introduction}
\end{figure*}


Technically, to capture the seasonality and trend within data, decomposition-based techniques are widely applied to disentangle the seasonality and trend parts explicitly. For example, DLinear and DUET apply the moving-average technique to obtain the trend part. TimesNet, PDF and TimeMixer apply meticulously-designed seasonal decomposition modules to process the seasonal part, and the CycleNet uses a learnable matrix to directly capture the seasonality. All these techniques are employed in the forward propagation to effectively extract the seasonal and trend components from the contextual time series. 

However, if the purpose of extracting seasonality and trend in the contextual time series is to improve predictions, perhaps considering seasonality and trends directly in the forecasting horizon may further enhance prediction performance. As shown in Figure~\ref{fig: introduction}, we observe that current distance-based loss functions~(such as MSE) have the following limitations: 1)~they may make poor seasonal predictions; 2)~they may make poor trend predictions; 3)~they may make both poor seasonal and trend predictions. Even when decomposition techniques are applied in the forward propagation, the seasonality and trend within the forecasting horizon are not effectively modeled, indicating that the inductive bias is not well applied to the predictions.


Inspired by the above motivation, we manage to explicitly encourage the modeling of the seasonality and trend in the forecasting horizon to enhance the performance. Specifically, we propose a simple yet effective \textbf{\underline{D}}ecomposition-\textbf{\underline{B}}ased \textbf{Loss} function called \textbf{DBLoss}. This method involves using exponential moving averages~\citep{xPatch} to decompose the time series into seasonal and trend components within the forecasting horizon. It then calculates the loss for each of these components separately and combines them with appropriate weighting. As a general loss function, combining DBLoss with any deep learning forecasting model can lead to consistent improvement in performance, which is demonstrated on real-world datasets from multiple domains. The contributions are summarized as follows. 
\begin{itemize}[left=0.3cm]
\item We propose a simple yet effective loss function for time series forecasting, called DBLoss, which can refine the characterization and representation of time series through decomposition within the forecasting horizon. 

\item The proposed DBLoss is generally applicable to arbitrary deep neural networks with negligible cost. By introducing DBLoss into the baseline, we have achieved performance that generally surpasses the state-of-the-art on eight real-world datasets.  

\item We conduct extensive evaluations of DBLoss using quantitative analysis and qualitative visualizations to verify its effectiveness.

\end{itemize}

\section{Related works}

\subsection{Time Series Forecasting Methods}
Time series forecasting (TSF) predicts future observations based on historical observations. TSF methods are mainly categorized into four distinct approaches: (1) statistical learning-based methods, (2) machine learning-based methods, (3) deep learning-based methods, and (4) foundation methods. 

Early TSF methods primarily rely on statistical learning approaches such as ARIMA~\citep{box1970distribution}, ETS~\citep{hyndman2008forecasting}, and VAR~\citep{godahewa2021monash}. With advancements in machine learning, methods like XGBoost~\citep{chen2016xgboost}, Random Forests~\citep{breiman2001random}, and LightGBM~\citep{ke2017lightgbm} gain popularity for handling nonlinear patterns. However, these methods still require manual feature engineering and model design~\citep{ma2025mofo,wang2023accurate,wu2025k2vae}. Leveraging the representation learning of deep neural networks (DNNs)~\citep{huang2023crossgnn,miao2025parameter,wang2025effective}, many deep learning-based methods emerge. TimesNet~\citep{TimesNet} and SegRNN~\citep{lin2023segrnn} model time series as vector sequences, using CNNs or RNNs to capture temporal dependencies. Additionally, Transformer architectures, including DUET~\citep{DUET}, Informer~\citep{zhou2021informer}, FEDformer~\citep{Fedformer}, Triformer~\citep{Triformer}, and PatchTST~\citep{PatchTST}, capture complex relationships between time points more accurately, significantly improving forecasting performance. MLP-based methods, including SparseTSF~\citep{lin2024sparsetsf}, CycleNet~\citep{lincyclenet}, SRSNet~\citep{wu2025srsnet}, NLinear~\citep{DLinear}, and DLinear~\citep{DLinear}, adopt simpler architectures with fewer parameters but still achieve highly competitive forecasting accuracy. 

However, many of these methods struggle with generalization across domains due to their reliance on domain-specific data~\citep{FM4TS-Bench}. To address this, foundation methods are proposed, categorized into LLM-based methods and time series pre-trained methods. LLM-based methods~\citep{gpt4ts, time-llm, unitime, S2IP-LLM} leverage the strong representational capacity and sequential modeling capability of LLMs to capture complex patterns for time series modeling. Time series pre-trained methods~\citep{Timer, units, Moment, timesfm} focus on pre-training over multi-domain time series data, enabling the method to learn domain-agnostic features that are transferable across various applications. This strategy not only enhances performance on specific tasks but also provides greater flexibility when adapting to new datasets or scenarios.

\subsection{Loss Functions for Time Series Forecasting}

Recently, to enhance the training performance of time series forecasting models, researchers have introduced various novel loss functions. These loss functions can be broadly categorized into three types: shape-based losses, dependency-based losses, and patch-based structural losses. 

Shape-based losses aim to capture structural similarities between true values and predictions by tackling the issue of shape mismatch. For example, techniques based on Dynamic Time Warping (DTW), such as Soft-DTW~\citep{Soft-dtw} and DILATE~\citep{DILATE}, can achieve alignment even when time series undergo deformation. However, despite their excellent performance in improving shape alignment, the high computational complexity of these methods restricts their application in large-scale scenarios. Meanwhile, TILDE-Q~\citep{TILDE-Q} introduces transformation invariance, making it robust to amplitude shifts, phase changes, and scale differences, thus focusing more on similarity at the shape level. Dependency-based losses are dedicated to characterizing temporal correlations within the forecasting horizon. For instance, FreDF~\citep{FreDF} cleverly circumvents complex correlation modeling between labels by performing learning and prediction in the frequency domain. Furthermore, patch-based structural losses like PSLoss~\citep{PSLoss} incorporate patch-wise statistical properties into the loss function, enabling a more granular structural measurement of the data. Unlike the aforementioned loss functions, our proposed DBLoss refines the characterization and representation of time series through decomposition within the forecasting horizon, offering a novel perspective for the design of time series loss functions.

\section{DBLoss}
\begin{figure*}[!tbp]
    \centering
    \includegraphics[width=1\linewidth]{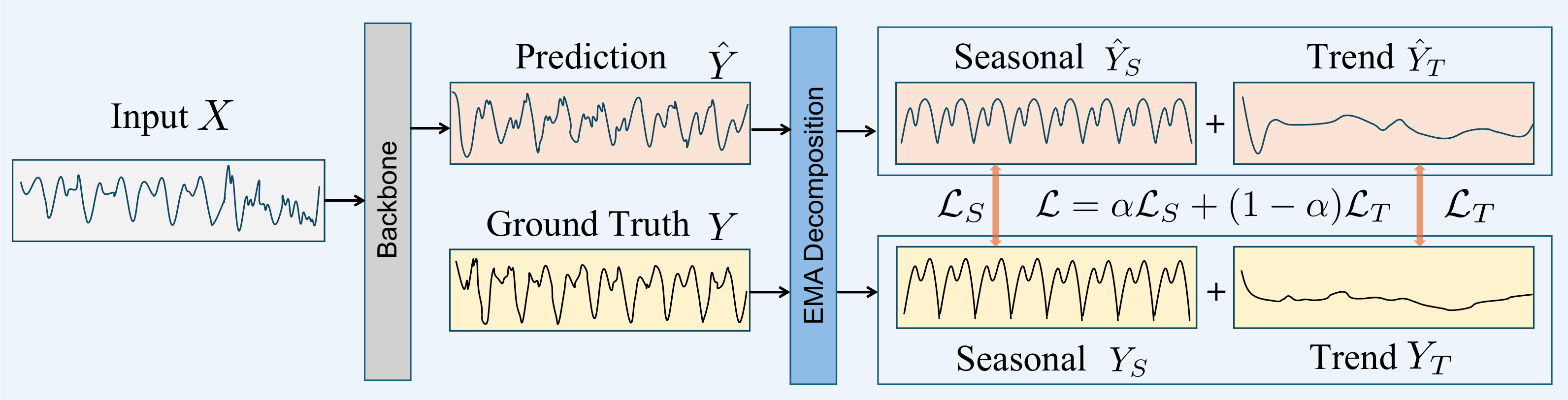}
    \caption{Overview of the proposed DBLoss.}
\label{fig: overview}
\end{figure*}

A \textit{time series} $\mX \in \mathbb{R}^{N\times T}$ is a time-oriented sequence of N-dimensional time points,  where $T$ is the number of timestamps, and $N$ is the number of channels. If $N=1$, a time series is called univariate, and multivariate if $N>1$. \textit{Time Series Forecasting} aims to predict the next $F$ future timestamps, formulated as $\mY =\langle \mX_{:,T+1}, \cdots, \mX_{:,T+F}\rangle~\in \mathbb{R}^{N \times F}$ based on the historical time series $\mX =~\langle \mX_{:,1}, \cdots, \mX_{:,T}\rangle~\in \mathbb{R}^{N \times T}$ with $N$ channels and $T$ timestamps. For convenience, we separate dimensions with commas. Specifically, we denote $\mX_{i,j} \in \mathbb{R}$ as the $i$-th channel at the $j$-th timestamp, $\mX_{n,:}\in \mathbb{R}^T$ as the time series of $n$-th channel, where $n=1,\cdots,N$. 

\subsection{Overview}
As shown in Figure \ref{fig: overview}, we first generate the prediction \(\hat{\mY}\) using an arbitrary backbone method. Next, we input both the prediction \(\hat{\mY}\) and the ground truth \(\mY\) into the EMA Decomposition Module to decompose them into seasonal and trend components. Through this process, we obtain the seasonal component \(\hat{\mY_S}\) and the trend component \(\hat{\mY_T}\) of the prediction, as well as the seasonal component \(\mY_S\) and the trend component \(\mY_T\) of the ground truth. Subsequently, we compute the errors for both the seasonal and trend components and then combine these errors using a weighted sum to form the final loss function. This approach allows for a more accuracy evaluation of the differences between the predicted and ground truth values, leading to more effective optimization and training.


\subsection{EMA Decomposition Module}

Seasonal-trend decomposition facilitates the learning of complex temporal patterns by breaking down time series signals into trend and seasonal components. Trend components refer to the long-term changes or patterns that occur over time, intuitively representing the overall direction of the data. In contrast, seasonal components capture the phenomena in the time series that repeat at specific intervals and are typically nonlinear due to the complexity and variability of periodic behavior. This technique is widely applied in time series analysis methods~\citep{Autoformer,Fedformer,DLinear,TimeMixer,DUET}. Unlike above methods, which typically extract the trend and seasonal representations of the time series through decomposition and then combine these two representations to obtain a more comprehensive time series representation for downstream tasks, our DBLoss computes the loss by separately decomposing the prediction and the ground truth into their trend and seasonal components. We then compute the losses for the trend and seasonal components separately and finally combine these losses using a weighted sum. This process enables the model to better capture the trends and seasonality of the ground truth, resulting in more accurate predictions.

There are various methods for seasonal-trend decomposition, such as STL decomposition~\citep{cleveland1990stl}, Simple Moving Average (SMA) decomposition~\citep{Autoformer,DUET,DLinear}, and Exponential Moving Average (EMA) decomposition~\citep{xPatch}. In this study, we chose EMA decomposition. Specifically, after obtaining the prediction \(\hat{\mY}\) and the ground truth \(\mY\), we input them into the EMA decomposition module to decompose them into their trend and seasonal components. We then compute the final loss in the weighted loss function described in Section \ref{Weighted Loss Function}.  Algorithm~\ref{alg1}
details the calculation process of EMA decomposition module.

\begin{algorithm}[h]
\caption{Calculation Process of EMA Decomposition Module}
\footnotesize
\begin{flushleft}
{\bf Input:} 
Time series $X \in \mathbb{R}^{B \times T \times N}$, where $B$ is the batch size, $T$ is the time steps, and $N$ is the number of channels; Smoothing factor $\alpha \in (0, 1)$

{\bf Output:}
Seasonality and Trend of $X$, denoted as $Seasonality \in \mathbb{R}^{B \times T \times N}, Trend \in \mathbb{R}^{B \times T \times N}$
\end{flushleft}
\begin{algorithmic}[1]
\State Get the shape of $X$: $B, T, N \leftarrow X.shape$
\State Calculate the weights: $W \leftarrow [(1-\alpha)^{T-1},(1-\alpha)^{T-2},\cdots, 1 ]$
\State Copy the weights to create a divisor: $D_{\text{div}} \leftarrow W.clone()$
\State Update the weights for EMA calculation: $W[1:] \leftarrow W[1:] \times \alpha$
\State Reshape the weights and divisor: 
\State \hspace{0.1in} $W \leftarrow W.reshape(1, T, 1)$
\State \hspace{0.1in} $D_{\text{div}} \leftarrow D_{\text{div}}.reshape(1, T, 1)$
\State Compute the cumulative sum of weighted data: $C \leftarrow \text{cumsum}(X \times W, \text{dim}=1)$
\State Divide the cumulative sum by the divisor: $Trend \leftarrow \frac{C}{D_{\text{div}}}$
\State Calculate the residual: $Seasonality \leftarrow X -Trend$
\State {\bf return} $Seasonality, Trend$
\end{algorithmic}
\label{alg1}
\end{algorithm}



\subsection{Weighted Loss Function}
\label{Weighted Loss Function}
Based on the EMA decomposition, we obtain the predicted seasonal component $\hat{Y_S}$ and trend component $\hat{Y_T}$, as well as the corresponding ground truth values $Y_S$ and $Y_T$. We then propose a weighted loss function, which consists of two parts: the seasonal loss $\mathcal{L}_S$ and the trend loss $\mathcal{L}_T$. 
\begin{align}
\mathcal{L_S} := \left|\hat{Y_S}-Y_S\right|_{2},  ~\mathcal{L_T} := \left|\hat{Y_T}-Y_T\right|_{1}.
\end{align}

To prevent the loss of one component from dominating the optimization process due to scale differences, we introduce a \textbf{scale alignment} mechanism.
Specifically, the trend loss is adaptively adjusted according to the relative magnitude between $\mathcal{L_S}$ and $\mathcal{L_T}$:
\begin{align}
\mathcal{L_T}^{\text{aligned}} 
:= \mathcal{L_T} \times 
\text{stopgrad}\left(\frac{\mathcal{L_S}}{\mathcal{L_T} + \epsilon}\right),
\end{align}
where $\epsilon$ is a small constant to ensure numerical stability. Here, $\text{stopgrad}(\cdot)$ denotes a \textbf{gradient detachment} operation, which prevents the gradient from back-propagating through the alignment ratio, thereby avoiding interference between the two loss components.

Finally, we define the total loss $\mathcal{L}$ as:
\begin{align}
\mathcal{L} := \beta \cdot \mathcal{L_S} + (1 - \beta) \cdot \mathcal{L_T}^{\text{aligned}},
\end{align}

where \(\beta\) is a tuning parameter used to balance the contributions of the seasonal loss and the trend loss. By adjusting the value of \(\beta\), we can optimize the model's training process according to specific application scenarios. 

We provide a theoretical analysis in Appendix~\ref{Theoretical Proofs} to explain why the proposed DBLoss is more effective than the conventional MSE loss for time series forecasting.


\section{Experiments}
\label{Experiments}
\subsection{Setup}
\label{Setup}
\textbf{Datasets} 
To conduct comprehensive and fair comparisons for different models, we conduct experiments on eight well-known forecasting benchmarks as the target datasets, including ETT (ETTh1, ETTh2, ETTm1, ETTm2), Solar, Weather, Electricity, and Traffic. For more details on the benchmark datasets, please refer to Table \(\ref{Multivariate datasets}\) in Appendix \(\ref{appendix dataset}\).

\textbf{Backbones}
We selected eight state-of-the-art (SOTA) time series forecasting models to serve as baselines. Specifically, we include four time series specific models: iTransformer~\citep{iTransformer}, Amplifier~\citep{Amplifier}, PatchTST~\citep{PatchTST}, and DLinear~\citep{DLinear}, as well as four time series foundation models: CALF~\citep{calf}, UniTS~\citep{units}, TTM~\citep{TTM}, and GPT4TS~\citep{gpt4ts}.

\textbf{Implementation Details}
To keep consistent with previous works, we adopt Mean Squared Error (MSE) and Mean Absolute Error (MAE) as evaluation metrics. We consider four forecasting horizon $F$: \{96, 192, 336, 720\} for all datasets. We utilize the comprehensive time series forecasting benchmark TFB~\citep{qiu2024tfb} for unified evaluation, with all baseline results also derived from TFB. \textit{Please note that for all the baseline scripts, we directly use the optimal scripts provided by TFB and only replace the training loss function with DBLoss, without making any other modifications}. The purpose of this approach is to validate the effectiveness of DBLoss to the greatest extent possible. By doing so, we can ensure the accuracy of the experimental results and clearly demonstrate the performance improvements brought by DBLoss. We do not apply the ``Drop Last'' trick~\citep{qiu2024tfb,qiu2025tab,qiu2025easytime} to ensure a fair comparison. All experiments of DBLoss are conducted using PyTorch in Python 3.8 and executed on an NVIDIA Tesla-A800 GPU. The training process is guided by the MSE loss function and employs the ADAM optimizer. The initial batch size is set to 64, with the flexibility to halve it (down to a minimum of 8) in case of an Out-Of-Memory (OOM) issue.

\subsection{Main results}

We present the MSE and MAE of four state-of-the-art long-term multivariate forecasting models on eight real-world datasets in Table~\ref{Long-term multivariate forecasting results}. Notably, DBLoss observes performance improvements across all backbone models and significantly outperforms MSE loss in most cases. This validates the robustness and broad applicability of the proposed loss function. Furthermore, DBLoss achieves significant improvements on models that have already adopted trend-seasonal decomposition to further extract better model representations, such as DLinear~\citep{DLinear}. This indicates that performing trend-seasonal decomposition during the loss computation does not conflict with the previous trend-seasonal decomposition operations but rather enhances model performance.


\begin{table*}[t]
\caption{Long-term multivariate forecasting results. The table reports MSE and MAE for different forecasting horizons $F \in \{96, 192, 336, 720\}$. The parameters for the baselines are kept consistent with those of TFB~\citep{qiu2024tfb}. The better results are highlighted in \textbf{bold}.}
\label{Long-term multivariate forecasting results}
\resizebox{\columnwidth}{!}{
\begin{tabular}{c|c|cccc|cccc|cccc|cccc}
    \toprule
\multicolumn{2}{c|}{Model} & \multicolumn{4}{c|}{iTransformer} & \multicolumn{4}{c|}{Amplifier} & \multicolumn{4}{c|}{PatchTST} & \multicolumn{4}{c}{DLinear} \\    \midrule
\multicolumn{2}{c|}{Loss} & \multicolumn{2}{c}{Ori} & \multicolumn{2}{c|}{DBLoss} & \multicolumn{2}{c}{Ori} & \multicolumn{2}{c|}{DBLoss} & \multicolumn{2}{c}{Ori} & \multicolumn{2}{c|}{DBLoss} & \multicolumn{2}{c}{Ori} & \multicolumn{2}{c}{DBLoss} \\     \midrule
\multicolumn{2}{c|}{Metric} & MSE & MAE & MSE & MAE & MSE & MAE & MSE & MAE & MSE & MAE & MSE & MAE & MSE & MAE & MSE & MAE \\ \midrule
 \multirow[c]{5}{*}{\rotatebox{90}{ETTh1}}& {96} & 0.386 & 0.405 & \textbf{0.383} & \textbf{0.396} & \textbf{0.376} & 0.393 & \textbf{0.376} & \textbf{0.389} & 0.377 & 0.397 & \textbf{0.373} & \textbf{0.390} & 0.379 & 0.403 & \textbf{0.369} & \textbf{0.390} \\
 & {192} & 0.424 & 0.440 & \textbf{0.405} & \textbf{0.421} & 0.414 & 0.42 & \textbf{0.409} & \textbf{0.415} & 0.409 & 0.425 & \textbf{0.395} & \textbf{0.413} & 0.408 & 0.419 & \textbf{0.402} & \textbf{0.409} \\
 & {336} & 0.449 & 0.460 & \textbf{0.425} & \textbf{0.438} & 0.442 & 0.446 & \textbf{0.430} & \textbf{0.432} & 0.431 & 0.444 & \textbf{0.414} & \textbf{0.426} & 0.440 & 0.440 & \textbf{0.430} & \textbf{0.428} \\
 & {720} & 0.495 & 0.487 & \textbf{0.478} & \textbf{0.463} & 0.48 & 0.479 & \textbf{0.459} & \textbf{0.465} & 0.457 & 0.477 & \textbf{0.425} & \textbf{0.451} & 0.471 & 0.493 & \textbf{0.449} & \textbf{0.475} \\\cmidrule{2-18} 
 & {Avg} & 0.439 & 0.448 & \textbf{0.423} & \textbf{0.430} & 0.428 & 0.435 & \textbf{0.419} & \textbf{0.425} & 0.419 & 0.436 & \textbf{0.402} & \textbf{0.420} & 0.425 & 0.439 & \textbf{0.412} & \textbf{0.425} \\\midrule
  \multirow[c]{5}{*}{\rotatebox{90}{ETTh2}}& {96} & 0.297 & 0.348 & \textbf{0.288} & \textbf{0.337} & 0.291 & 0.342 & \textbf{0.288} & \textbf{0.332} & \textbf{0.274} & 0.337 & \textbf{0.274} & \textbf{0.334} & 0.300 & 0.364 & \textbf{0.284} & \textbf{0.342} \\
 & {192} & 0.372 & 0.403 & \textbf{0.357} & \textbf{0.389} & 0.355 & 0.4 & \textbf{0.344} & \textbf{0.379} & 0.348 & 0.384 & \textbf{0.334} & \textbf{0.376} & 0.387 & 0.423 & \textbf{0.357} & \textbf{0.390} \\
 & {336} & 0.388 & 0.417 & \textbf{0.385} & \textbf{0.416} & 0.384 & 0.42 & \textbf{0.377} & \textbf{0.405} & 0.377 & 0.416 & \textbf{0.349} & \textbf{0.392} & 0.490 & 0.487 & \textbf{0.407} & \textbf{0.430} \\
 & {720} & \textbf{0.424} & 0.444 & 0.427 & \textbf{0.443} & 0.422 & 0.451 & \textbf{0.400} & \textbf{0.437} & 0.406 & 0.441 & \textbf{0.390} & \textbf{0.422} & 0.704 & 0.597 & \textbf{0.586} & \textbf{0.533} \\\cmidrule{2-18} 
 & {Avg} & 0.370 & 0.403 & \textbf{0.364} & \textbf{0.396} & 0.363 & 0.403 & \textbf{0.352} & \textbf{0.388} & 0.351 & 0.395 & \textbf{0.337} & \textbf{0.381} & 0.470 & 0.468 & \textbf{0.409} & \textbf{0.424} \\\midrule
  \multirow[c]{5}{*}{\rotatebox{90}{ETTm1}}& {96} & 0.300 & 0.353 & \textbf{0.290} & \textbf{0.341} & 0.293 & 0.347 & \textbf{0.287} & \textbf{0.335} & 0.289 & 0.343 & \textbf{0.284} & \textbf{0.328} & 0.300 & 0.345 & \textbf{0.295} & \textbf{0.337} \\
 & {192} & 0.341 & 0.380 & \textbf{0.328} & \textbf{0.363} & 0.329 & 0.367 & \textbf{0.328} & \textbf{0.359} & 0.329 & 0.368 & \textbf{0.322} & \textbf{0.355} & 0.336 & 0.366 & \textbf{0.331} & \textbf{0.358} \\
 & {336} & 0.374 & 0.396 & \textbf{0.368} & \textbf{0.386} & 0.365 & 0.387 & \textbf{0.364} & \textbf{0.380} & 0.362 & 0.390 & \textbf{0.359} & \textbf{0.376} & 0.367 & 0.386 & \textbf{0.361} & \textbf{0.378} \\
 & {720} & 0.429 & 0.430 & \textbf{0.415} & \textbf{0.415} & 0.429 & 0.422 & \textbf{0.424} & \textbf{0.413} & 0.416 & 0.423 & \textbf{0.410} & \textbf{0.412} & 0.419 & 0.416 & \textbf{0.415} & \textbf{0.409} \\\cmidrule{2-18} 
 & {Avg} & 0.361 & 0.390 & \textbf{0.350} & \textbf{0.376} & 0.354 & 0.381 & \textbf{0.351} & \textbf{0.372} & 0.349 & 0.381 & \textbf{0.344} & \textbf{0.368} & 0.356 & 0.378 & \textbf{0.351} & \textbf{0.370} \\\midrule
  \multirow[c]{5}{*}{\rotatebox{90}{ETTm2}}& {96} & 0.175 & 0.266 & \textbf{0.166} & \textbf{0.254} & 0.168 & 0.258 & \textbf{0.163} & \textbf{0.245} & 0.165 & 0.255 & \textbf{0.163} & \textbf{0.246} & 0.164 & 0.255 & \textbf{0.163} & \textbf{0.247} \\
 & {192} & 0.242 & 0.312 & \textbf{0.227} & \textbf{0.295} & 0.227 & 0.298 & \textbf{0.222} & \textbf{0.288} & 0.221 & 0.293 & \textbf{0.219} & \textbf{0.284} & 0.224 & 0.304 & \textbf{0.220} & \textbf{0.290} \\
 & {336} & 0.282 & 0.337 & \textbf{0.278} & \textbf{0.330} & 0.276 & 0.334 & \textbf{0.271} & \textbf{0.322} & 0.276 & 0.327 & \textbf{0.273} & \textbf{0.320} & \textbf{0.277} & 0.337 & \textbf{0.277} & \textbf{0.329} \\
 & {720} & \textbf{0.375} & 0.394 & \textbf{0.375} & \textbf{0.388} & 0.364 & 0.394 & \textbf{0.350} & \textbf{0.373} & 0.362 & 0.381 & \textbf{0.357} & \textbf{0.374} & 0.371 & 0.401 & \textbf{0.366} & \textbf{0.390} \\\cmidrule{2-18} 
 & {Avg} & 0.269 & 0.327 & \textbf{0.262} & \textbf{0.317} & 0.259 & 0.321 & \textbf{0.252} & \textbf{0.307} & 0.256 & 0.314 & \textbf{0.253} & \textbf{0.306} & 0.259 & 0.324 & \textbf{0.257} & \textbf{0.314} \\\midrule
  \multirow[c]{5}{*}{\rotatebox{90}{Solar}} & {96} & 0.190 & 0.244 & \textbf{0.180} & \textbf{0.215} & \textbf{0.184} & 0.239 & 0.189 & \textbf{0.226} & 0.170 & 0.234 & \textbf{0.167} & \textbf{0.211} & \textbf{0.199} & 0.265 & 0.202 & \textbf{0.236} \\
 & {192} & \textbf{0.193} & 0.257 & 0.201 & \textbf{0.239} & \textbf{0.202} & 0.252 & 0.208 & \textbf{0.239} & 0.204 & 0.302 & \textbf{0.182} & \textbf{0.226} & \textbf{0.220} & 0.282 & 0.224 & \textbf{0.250} \\
 & {336} & 0.203 & 0.266 & \textbf{0.195} & \textbf{0.232} & \textbf{0.232} & 0.274 & 0.235 & \textbf{0.251} & 0.212 & 0.293 & \textbf{0.187} & \textbf{0.232} & \textbf{0.234} & 0.295 & 0.237 & \textbf{0.256} \\
 & {720} & \textbf{0.223} & 0.281 & 0.232 & \textbf{0.265} & \textbf{0.229} & 0.276 & 0.242 & \textbf{0.256} & 0.215 & 0.307 & \textbf{0.197} & \textbf{0.237} & \textbf{0.243} & 0.301 & 0.245 & \textbf{0.260} \\\cmidrule{2-18} 
 & {Avg} & \textbf{0.202} & 0.262 & \textbf{0.202} & \textbf{0.238} & \textbf{0.212} & 0.260 & 0.219 & \textbf{0.243} & 0.200 & 0.284 & \textbf{0.183} & \textbf{0.227} & \textbf{0.224} & 0.286 & 0.227 & \textbf{0.251} \\\midrule
\multirow[c]{5}{*}{\rotatebox{90}{Weather}} & {96} & 0.157 & 0.207 & \textbf{0.154} & \textbf{0.196} & 0.147 & 0.199 & \textbf{0.145} & \textbf{0.189} & 0.150 & 0.200 & \textbf{0.149} & \textbf{0.189} & 0.170 & 0.230 & \textbf{0.169} & \textbf{0.221} \\
 & {192} & 0.200 & 0.248 & \textbf{0.197} & \textbf{0.239} & 0.188 & 0.238 & \textbf{0.186} & \textbf{0.228} & 0.191 & 0.239 & \textbf{0.189} & \textbf{0.229} & \textbf{0.216} & 0.273 & \textbf{0.216} & \textbf{0.262} \\
 & {336} & 0.252 & 0.287 & \textbf{0.249} & \textbf{0.278} & \textbf{0.239} & 0.276 & \textbf{0.239} & \textbf{0.269} & 0.242 & 0.279 & \textbf{0.240} & \textbf{0.270} & 0.258 & 0.307 & \textbf{0.253} & \textbf{0.293} \\
 & {720} & 0.320 & 0.336 & \textbf{0.319} & \textbf{0.335} & \textbf{0.316} & 0.328 & \textbf{0.316} & \textbf{0.323} & \textbf{0.312} & 0.330 & 0.314 & \textbf{0.322} & 0.323 & 0.362 & \textbf{0.319} & \textbf{0.346} \\\cmidrule{2-18} 
& {Avg} & 0.232 & 0.270 & \textbf{0.230} & \textbf{0.262} & 0.223 & 0.260 & \textbf{0.221} & \textbf{0.252} & 0.224 & 0.262 & \textbf{0.223} & \textbf{0.252} & 0.242 & 0.293 & \textbf{0.239} & \textbf{0.280} \\\midrule
\multirow[c]{5}{*}{\rotatebox{90}{Electricity}} & {96} & 0.134 & 0.230 & \textbf{0.131} & \textbf{0.226} & \textbf{0.132} & \textbf{0.227} & 0.133 & \textbf{0.227} & \textbf{0.143} & 0.247 & \textbf{0.143} & \textbf{0.244} & \textbf{0.140} & 0.237 & \textbf{0.140} & \textbf{0.235} \\
 & {192} & 0.154 & 0.250 & \textbf{0.149} & \textbf{0.242} & 0.149 & 0.241 & \textbf{0.147} & \textbf{0.239} & \textbf{0.158} & 0.260 & \textbf{0.158} & \textbf{0.257} & \textbf{0.154} & 0.251 & \textbf{0.154} & \textbf{0.247} \\
 & {336} & 0.169 & 0.265 & \textbf{0.163} & \textbf{0.257} & 0.165 & 0.258 & \textbf{0.163} & \textbf{0.256} & 0.168 & 0.267 & \textbf{0.165} & \textbf{0.259} & \textbf{0.169} & 0.268 & \textbf{0.169} & \textbf{0.264} \\
 & {720} & \textbf{0.194} & 0.288 & 0.195 & \textbf{0.284} & \textbf{0.203} & 0.292 & \textbf{0.203} & \textbf{0.290} & \textbf{0.214} & 0.307 & \textbf{0.214} & \textbf{0.304} & 0.204 & 0.301 & \textbf{0.203} & \textbf{0.295} \\\cmidrule{2-18} 
& {Avg} & 0.163 & 0.258 & \textbf{0.160} & \textbf{0.252} & \textbf{0.162} & 0.255 & \textbf{0.162} & \textbf{0.253} & 0.171 & 0.270 & \textbf{0.170} & \textbf{0.266} & \textbf{0.167} & 0.264 & \textbf{0.167} & \textbf{0.260} \\\midrule
\multirow[c]{5}{*}{\rotatebox{90}{Traffic}}  & {96} & \textbf{0.363} & 0.265 & 0.366 & \textbf{0.261} & 0.396 & 0.278 & \textbf{0.393} & \textbf{0.270} & 0.370 & 0.262 & \textbf{0.369} & \textbf{0.254} & \textbf{0.395} & 0.275 & 0.396 & \textbf{0.270} \\
 & {192} & \textbf{0.384} & 0.273 & 0.387 & \textbf{0.271} & 0.413 & 0.285 & \textbf{0.412} & \textbf{0.275} & 0.386 & 0.269 & \textbf{0.385} & \textbf{0.260} & \textbf{0.407} & 0.280 & \textbf{0.407} & \textbf{0.274} \\
 & {336} & \textbf{0.396} & 0.277 & 0.397 & \textbf{0.275} & \textbf{0.421} & 0.291 & 0.422 & \textbf{0.286} & 0.396 & 0.275 & \textbf{0.395} & \textbf{0.266} & 0.417 & 0.286 & \textbf{0.415} & \textbf{0.279} \\
 & {720} & 0.445 & 0.308 & \textbf{0.444} & \textbf{0.306} & \textbf{0.456} & 0.307 & \textbf{0.456} & \textbf{0.304} & 0.435 & 0.295 & \textbf{0.432} & \textbf{0.286} & 0.454 & 0.308 & \textbf{0.449} & \textbf{0.298} \\\cmidrule{2-18} 
& {Avg} & \textbf{0.397} & 0.281 & 0.399 & \textbf{0.278} & 0.422 & 0.290 & \textbf{0.421} & \textbf{0.284} & 0.397 & 0.275 & \textbf{0.395} & \textbf{0.267} & 0.418 & 0.287 & \textbf{0.417} & \textbf{0.280}\\
\bottomrule
\end{tabular}}
\end{table*}

\subsection{Comparison with Other Loss Functions}
To better validate the effectiveness of DBLoss, we compare it with several other loss functions---see Table~\ref{Comparison between the proposed DBLoss and other loss functions}. TILDE-Q emphasizes shape similarity using transformation-invariant loss terms. FreDF cleverly circumvents complex correlation modeling between labels by performing learning and prediction in the frequency domain. PSLoss incorporates patch-wise statistical properties into the loss function, enabling a more granular structural measurement of the data. The results indicate that DBLoss achieves the lowest MSE and MAE in most cases across various datasets and forecasting horizons. This is due to its ability to refine the characterization and representation of time series through decomposition within the forecasting horizon, thereby achieving a more precise alignment between the ground truth and predictions.

\begin{table*}[!t]
  \centering
  \caption{Comparison between the proposed DBLoss and other loss functions. The model is DLinear and we report the result of three datasets-ETTh2, ETTm1, and Traffic. The best results are highlighted in \textbf{bold}, and the second-best results are highlighted in \underline{underline}.}
  \label{Comparison between the proposed DBLoss and other loss functions}
  \resizebox{1\columnwidth}{!}{%
    \begin{tabular}{c|c|ccccc|ccccc|ccccc}
    \toprule
    \multicolumn{2}{c|}{Dataset} & \multicolumn{5}{c|}{ETTh2}    & \multicolumn{5}{c|}{ETTm1}    & \multicolumn{5}{c}{Traffic} \\
    \midrule
    \multicolumn{2}{c|}{Forecast horizon} & 96    & 192   & 336   & 720  & Avg & 96    & 192   & 336   & 720 &  Avg & 96    & 192   & 336   & 720&  Avg \\
    \midrule
    \multirow{2}[2]{*}{Ori} & MSE   & 0.300 & 0.387 & 0.490 & 0.704 &0.470& 0.300 & 0.336 & 0.367 & 0.419 & 0.356 &\textbf{0.395} & \textbf{0.407} & 0.417 & 0.454 & \underline{0.418}\\
          & MAE   & 0.364 & 0.423 & 0.487 & 0.597 & 0.468 & 0.345 & 0.366 & 0.386 & 0.416 & 0.378& 0.275 & 0.280 & 0.286 &0.308& 0.287\\
    \midrule
    TILDE-Q & MSE   & 0.287 & 0.362 & 0.425 & 0.599 & 0.418 & 0.302 & 0.336 & 0.371 & 0.425 & 0.359 & 0.416 & 0.422 &0.423& 0.461 &0.431\\
    (\citeyear{TILDE-Q}) & MAE   & 0.345 & 0.395 & 0.445 & 0.551 & 0.434 & 0.342 & 0.362 & 0.386 & 0.417 & 0.377 & 0.294 & 0.296 &0.293&0.316&0.300\\
    \midrule
    FreDF & MSE   & \underline{0.284} & 0.362 & 0.420 &  \underline{0.587} &  \underline{0.413} & 0.302& 0.333 & 0.363 & \textbf{0.415}& 0.353 & 0.398 &0.408 & \underline{0.416}& \underline{0.452}& 0.419\\
    (\citeyear{FreDF}) & MAE   & \textbf{0.342} & 0.396& 0.445&  \underline{0.546} &0.432 & 0.344 & 0.363& 0.381 &  \underline{0.411}& 0.375 &\textbf{0.270} & 0.275 &0.280&0.302&0.282\\
    \midrule
    PSLoss & MSE   & \textbf{0.283}& \underline{0.358} & \underline{0.411} & 0.614& 0.417 &  \underline{0.296} &  \underline{0.332}& \textbf{0.361} & 0.416 &  \textbf{0.351} & 0.398 & 0.408 & \underline{0.416}& \underline{0.452}&0.419\\
    (\citeyear{PSLoss}) & MAE   & 0.343 & \underline{0.393} & \underline{0.434}& 0.549 &  \underline{0.430} &  \underline{0.339} &  \underline{0.361} &  \underline{0.380}& 0.413 &  \underline{0.373}& \textbf{0.270} & \textbf{0.274} &\textbf{0.279}& \underline{0.299}&  \underline{0.281}\\
    \midrule
   DBLoss & MSE   & \underline{0.284 } & \textbf{0.357 } & \textbf{0.407 }& \textbf{0.586 } & \textbf{0.409}& \textbf{0.295 } & \textbf{0.331} & \textbf{0.361} & \textbf{0.415} & \textbf{0.351} & \underline{0.396}&\textbf{0.407}&\textbf{0.415}&\textbf{0.449}& \textbf{0.417}\\
        (Ours)  & MAE   & \textbf{0.342}& \textbf{0.390}& \textbf{0.430} & \textbf{0.533}& \textbf{0.424} & \textbf{0.337} & \textbf{0.358} & \textbf{0.378}& \textbf{0.409}& \textbf{0.370} & \textbf{0.270} & \textbf{0.274} &\textbf{0.279}&\textbf{0.298}& \textbf{0.280}\\
    \bottomrule
    \end{tabular}%
    }
  \label{table:loss compare}%
\end{table*}%

\subsection{Zero-shot Forecasting Results}
To evaluate the effectiveness of DBLoss in enhancing the generalization ability on unseen datasets, we follow the methods outlined in \citep{chen2024similarity,PSLoss} and conduct zero-shot forecasting experiments. Specifically, we sequentially use ETTh1, ETTh2, ETTm1, and ETTm2 as source datasets, while the remaining datasets serve as target datasets. 

Table~\ref{tab:zero-shot} shows the results measured on the target datasets when the forecasting horizon is set to 720. These results highlight the consistent advantages of DBLoss. In most cases, DBLoss outperforms MSE loss, indicating that it can significantly improve the model's generalization performance across different datasets and sampling frequencies. These improvements stem from DBLoss's ability to better capture the intrinsic trends and seasonal patterns within the datasets, thereby enabling the model to more effectively adapt to unseen data patterns.

\begin{table}[t!]
\centering
\caption{Zero-shot forecasting results on ETT datasets. The forecasting horizon is 720. The parameters for the baselines are kept consistent with those of TFB~\citep{qiu2024tfb}. The better results are highlighted in \textbf{bold}.}
\label{tab:zero-shot}
\resizebox{\linewidth}{!}{
\begin{tabular}{c|cccc|cccc|cccc|cccc}
    \toprule
Model & \multicolumn{4}{c|}{iTransformer} & \multicolumn{4}{c|}{Amplifier} & \multicolumn{4}{c|}{PatchTST} & \multicolumn{4}{c}{DLinear} \\    \midrule
Loss & \multicolumn{2}{c}{Ori} & \multicolumn{2}{c|}{DBLoss} & \multicolumn{2}{c}{Ori} & \multicolumn{2}{c|}{DBLoss} & \multicolumn{2}{c}{Ori} & \multicolumn{2}{c|}{DBLoss} & \multicolumn{2}{c}{Ori} & \multicolumn{2}{c}{DBLoss} \\     \midrule
Metric & MSE & MAE & MSE & MAE & MSE & MAE & MSE & MAE & MSE & MAE & MSE & MAE & MSE & MAE & MSE & MAE \\ \midrule
ETTh1$\rightarrow$ETTh2 &0.461&0.470&\textbf{0.434}&\textbf{0.446}&\textbf{0.393}&\textbf{0.427}&0.401&0.431&0.402&0.437&\textbf{0.389}& \textbf{0.427}&0.647&0.573&\textbf{0.542}&\textbf{0.520} \\\cmidrule{2-17} 
  ETTh1$\rightarrow$ETTm1 &1.061&0.676&\textbf{0.771}&\textbf{0.592}&0.777&\textbf{0.571}&\textbf{0.758}&0.576&0.753&0.590&\textbf{0.722}&\textbf{0.570}& 0.754&0.602 &\textbf{0.735}&\textbf{0.584}\\ \cmidrule{2-17} 
 ETTh1$\rightarrow$ETTm2 &0.454&0.447&\textbf{0.434}&\textbf{0.420}&\textbf{0.406}&0.415&0.412&\textbf{0.414}&0.403&0.414&\textbf{0.400}&\textbf{0.410}& 0.640& 0.566 &\textbf{0.535}&\textbf{0.510}\\ \midrule
ETTh2$\rightarrow$ETTh1 &0.672&0.593&\textbf{0.557}&\textbf{0.521}&0.678&0.592&\textbf{0.530}&\textbf{0.515}&0.593&0.556&\textbf{0.484}&\textbf{0.490}& 0.506&0.521&\textbf{0.450}&\textbf{0.477}\\ \cmidrule{2-17} 
 ETTh2$\rightarrow$ETTm1 &0.969&0.659&\textbf{0.802}&\textbf{0.594}&0.761&0.585&\textbf{0.714}&\textbf{0.564}&0.762&0.577&\textbf{0.738}&\textbf{0.551}& 0.752&0.608&\textbf{0.738}&\textbf{0.579}\\ \cmidrule{2-17} 
ETTh2$\rightarrow$ETTm2 &\textbf{0.417}&0.428&0.436&\textbf{0.422}&0.403&0.417&\textbf{0.403}&\textbf{0.415}&\textbf{0.393}&0.409&0.395&\textbf{0.404}& 0.787&0.629&\textbf{0.580}&\textbf{0.526}\\ \midrule
 ETTm1$\rightarrow$ETTh1 &0.705&0.598&\textbf{0.528}&\textbf{0.516}&0.500&0.494&\textbf{0.482}&\textbf{0.488}&0.710&0.594&\textbf{0.553}&\textbf{0.534}& 0.460&0.481&\textbf{0.445}&\textbf{0.468}\\ \cmidrule{2-17} 
ETTm1$\rightarrow$ETTh2 &0.433&0.460&\textbf{0.409}&\textbf{0.444}&0.425&0.446&\textbf{0.421}&\textbf{0.445}&\textbf{0.418}&\textbf{0.451}&0.431&0.452& 0.427&0.464&\textbf{0.404}&\textbf{0.444}\\ \cmidrule{2-17} 
 ETTm1$\rightarrow$ETTm2&\textbf{0.369}&0.389&0.370&\textbf{0.387}&\textbf{0.369}&0.384&0.372&\textbf{0.384}&0.370&0.391&\textbf{0.367}&\textbf{0.384}& 0.389&0.416&\textbf{0.367}&\textbf{0.394}\\ \midrule
ETTm2$\rightarrow$ETTh1 &1.001&0.704&\textbf{0.775}&\textbf{0.613}&0.542&0.524&\textbf{0.479}&\textbf{0.491}&0.896&0.695&\textbf{0.617}&\textbf{0.577}& 0.488&0.497&\textbf{0.460}&\textbf{0.481}\\ \cmidrule{2-17} 
 ETTm2$\rightarrow$ETTh2 &0.477&0.486&\textbf{0.456}&\textbf{0.468}&0.444&0.464&\textbf{0.414}&\textbf{0.439}&0.412&0.449&\textbf{0.400}&\textbf{0.431} & 0.415&0.452&\textbf{0.410}&\textbf{0.445}\\ \cmidrule{2-17} 
ETTm2$\rightarrow$ETTm1 &0.662&0.566&\textbf{0.551}&\textbf{0.498}&0.652&0.547&\textbf{0.478}&\textbf{0.452}&0.484&0.451&\textbf{0.452}&\textbf{0.429}&0.449&0.439&\textbf{0.436}& \textbf{0.430}\\
 \bottomrule
\end{tabular}}
\end{table}

\subsection{Results on Time Series Foundation Models}
To further evaluate the effectiveness of the proposed DBLoss, we conducted 5\% few-shot experiments on four time series foundation models, using DBLoss only during the fine-tuning stage. These models include two LLM-based time series forecasting models: CALF~\citep{calf}, GPT4TS~\citep{gpt4ts}, as well as two time series 
 pre-trained models: UniTS~\citep{units}, TTM~\citep{TTM}. The results in Table~\ref{Foundation models long-term multivariate forecasting results} show that incorporating DBLoss consistently outperforms the standard MSE loss. These findings highlight that DBLoss not only enhances performance on specific models but also improves the performance of foundation models, further demonstrating its significant role in multivariate time series forecasting.

\begin{table*}[t]
\caption{Foundation models results in the 5\% few-shot setting. The table reports average MSE and MAE for four forecasting lengths $F \in \{96, 192, 336, 720\}$. The parameters for the baselines are kept consistent with those of TSFM-Bench~\citep{li2025TSFM-Bench}. The better results are highlighted in \textbf{bold}. Full results are provided in Table~\ref{Foundation models} of Appendix~\ref{Full Results on Time Series Foundation Models}}
\label{Foundation models long-term multivariate forecasting results}
\resizebox{\columnwidth}{!}{
\begin{tabular}{c|cccc|cccc|cccc|cccc}
\toprule
\multicolumn{1}{c|}{Model} & \multicolumn{4}{c|}{GPT4TS} & \multicolumn{4}{c|}{CALF} & \multicolumn{4}{c|}{TTM} & \multicolumn{4}{c}{UniTS} \\\midrule
\multicolumn{1}{c|}{Loss} & \multicolumn{2}{c}{Ori} & \multicolumn{2}{c|}{DBLoss} & \multicolumn{2}{c}{Ori} & \multicolumn{2}{c|}{DBLoss} & \multicolumn{2}{c}{Ori} & \multicolumn{2}{c|}{DBLoss} & \multicolumn{2}{c}{Ori} & \multicolumn{2}{c}{DBLoss} \\     \midrule
\multicolumn{1}{c|}{Metric} & MSE & MAE & MSE & MAE & MSE & MAE & MSE & MAE & MSE & MAE & MSE & MAE & MSE & MAE & MSE & MAE \\ \midrule
{ETTh1} & 0.467 & 0.470 & \textbf{0.453} & \textbf{0.462} & 0.443 & 0.454 & \textbf{0.433} & \textbf{0.446} & 0.405 & 0.425 & \textbf{0.395} & \textbf{0.417} & 0.436 & 0.434 & \textbf{0.425} & \textbf{0.427} \\\midrule
{ETTh2} & 0.373 & 0.414 & \textbf{0.368} & \textbf{0.406} & 0.373 & 0.407 & \textbf{0.368} & \textbf{0.404} & 0.342 & 0.383 & \textbf{0.332} & \textbf{0.378} & 0.372 & 0.405 & \textbf{0.357} & \textbf{0.393} \\\midrule
{ETTm1} & 0.388 & 0.404 & \textbf{0.377} & \textbf{0.394} & 0.372 & 0.396 & \textbf{0.358} & \textbf{0.382} & 0.356 & 0.376 & \textbf{0.354} & \textbf{0.372} & 0.377 & 0.402 & \textbf{0.362} & \textbf{0.386} \\\midrule
{ETTm2} & 0.278 & 0.335 & \textbf{0.266} & \textbf{0.320} & 0.271 & 0.332 & \textbf{0.259} & \textbf{0.316} & 0.258 & 0.313 & \textbf{0.257} & \textbf{0.308} & 0.292 & 0.344 & \textbf{0.270} & \textbf{0.320} \\\midrule
{Solar} & 0.262 & 0.335 & \textbf{0.254} & \textbf{0.279} & \textbf{0.229} & \textbf{0.297} & 0.246 & 0.300 & \textbf{0.219} & 0.269 & 0.224 & \textbf{0.266} & \textbf{0.206} & 0.261 & 0.214 & \textbf{0.246} \\\midrule
{Weather} & 0.253 & 0.293 & \textbf{0.248} & \textbf{0.284} & 0.238 & 0.277 & \textbf{0.236} & \textbf{0.272} & \textbf{0.225} & 0.260 & \textbf{0.225} & \textbf{0.256} & \textbf{0.230} & 0.269 & 0.231 & \textbf{0.260} \\\midrule
{Electricity} & \textbf{0.207} & 0.317 & \textbf{0.207} & \textbf{0.309} & 0.172 & 0.268 & \textbf{0.171} & \textbf{0.264} & 0.179 & 0.277 & \textbf{0.178} & \textbf{0.274} & \textbf{0.180} & 0.275 & 0.181 & \textbf{0.274} \\\midrule
{Traffic} & 0.433 & 0.309 & \textbf{0.428} & \textbf{0.295} & 0.435 & 0.316 & \textbf{0.433} & \textbf{0.309} & 0.484 & 0.341 & \textbf{0.481} & \textbf{0.339} & 0.422 & 0.289 & \textbf{0.420} & \textbf{0.282}\\ \bottomrule
\end{tabular}}
\end{table*}




\subsection{Impact of DBLoss on Generalization}
\begin{figure*}[!htbp]
  \centering
  \subfloat[Results on ETTh1 dataset]
  {\includegraphics[width=0.495\textwidth]{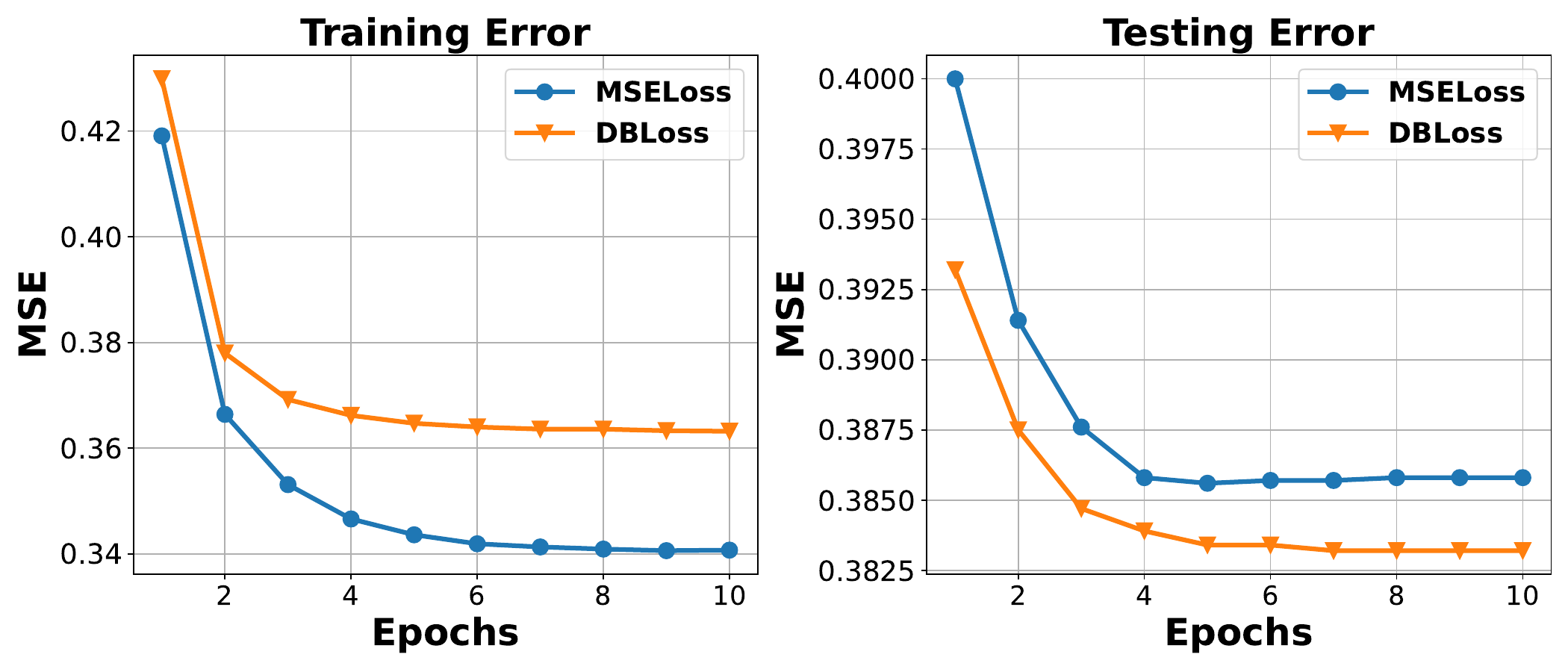}\label{Results on ETTh1 dataset}}
  \hspace{0.5mm}\subfloat[Results on ETTm2 dataset]
  {\includegraphics[width=0.495\textwidth]{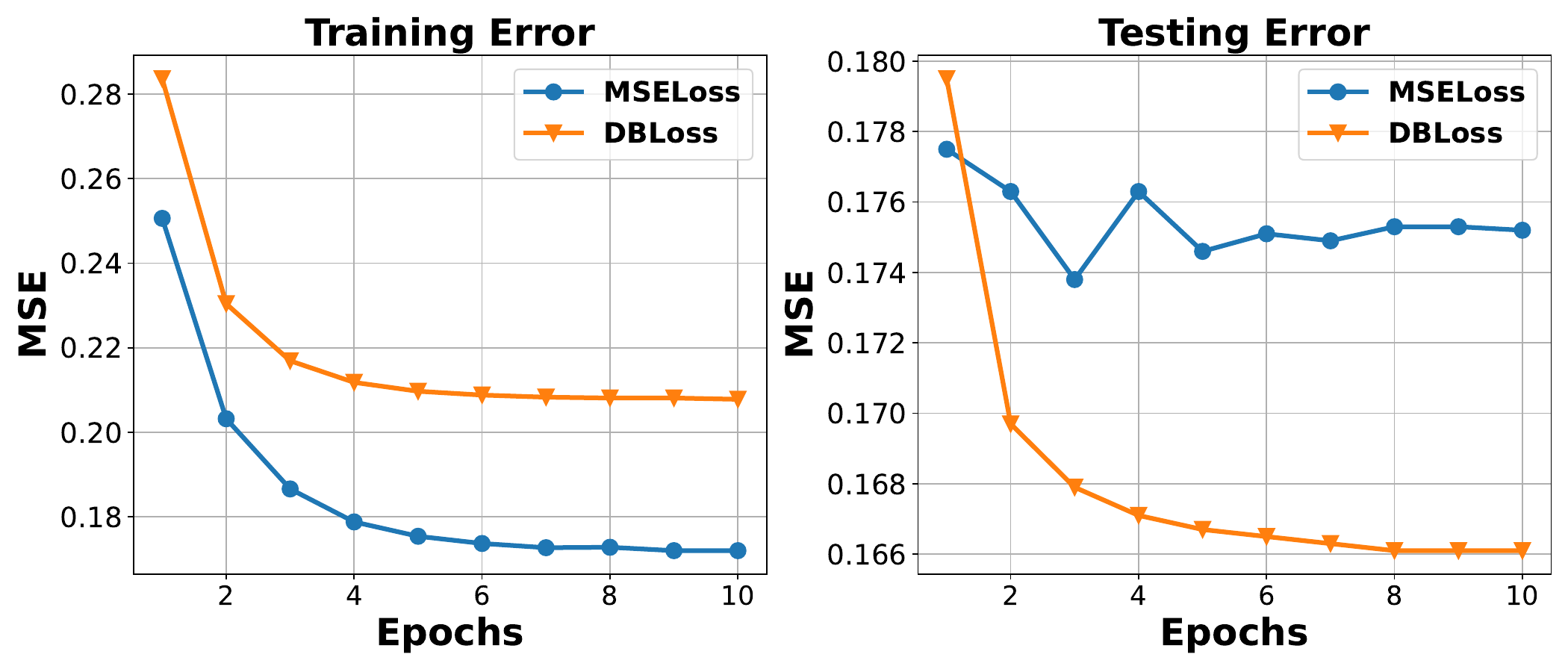}\label{Results on ETTm2 dataset}}

  \hspace{0.5mm}\subfloat[Results on Weather dataset]
  {\includegraphics[width=0.495\textwidth]{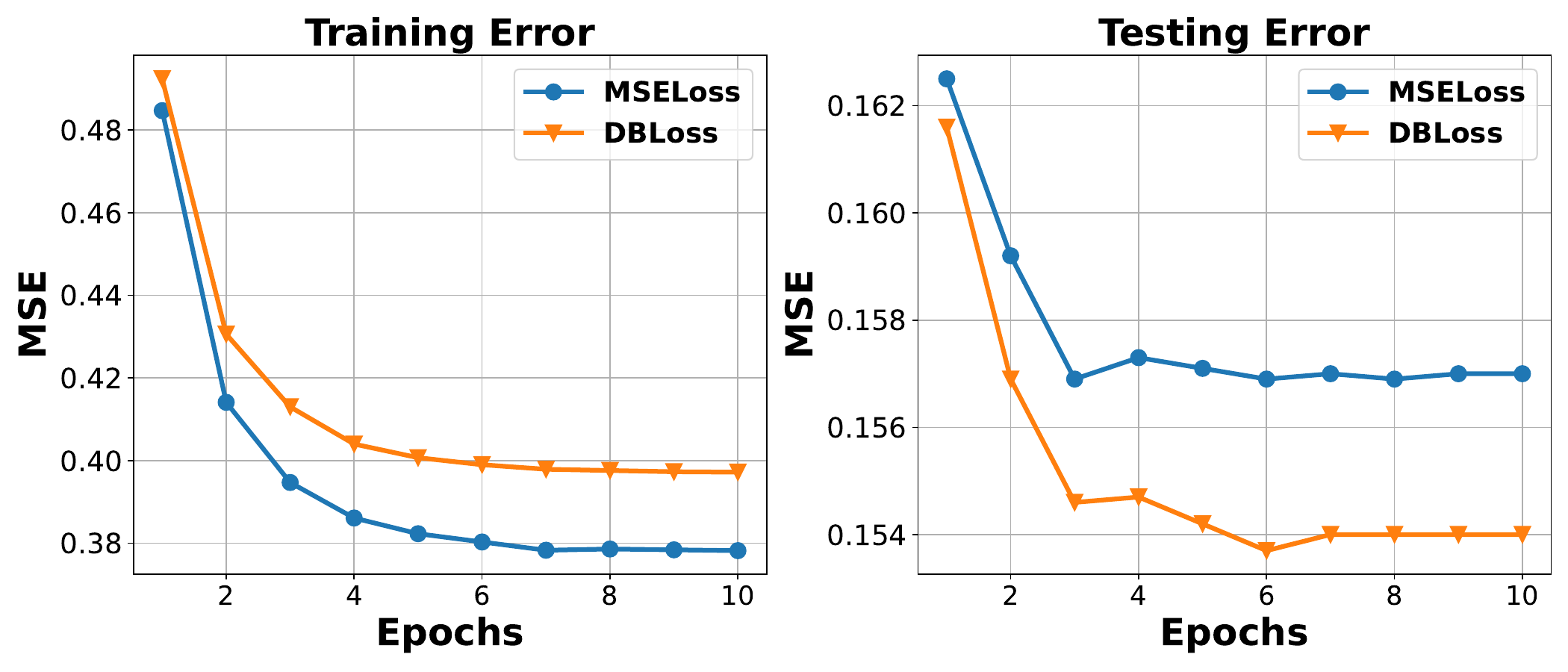}\label{Results on Weather dataset}}
 \hspace{0.5mm}\subfloat[Results on Solar dataset]
  {\includegraphics[width=0.495\textwidth]{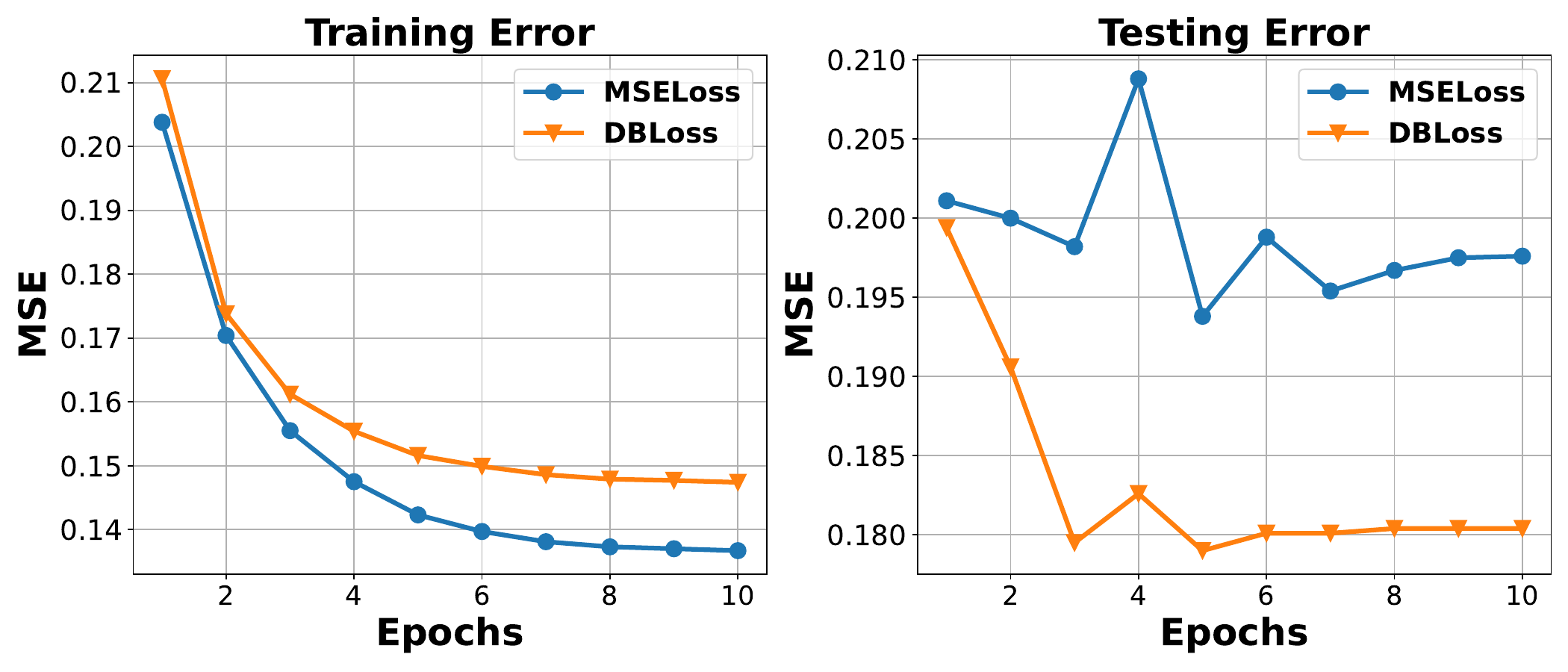}\label{Results on Solar dataset}}
  \caption{Training and testing MSE loss curves across all training epochs for the iTransformer model trained with MSE loss and DBLoss on the ETTh1, ETTm2, Weather, and Solar datasets. Notably, the model trained with DBLoss exhibits higher training errors but achieves lower testing errors. This highlights the effectiveness of DBLoss in enhancing generalization and mitigating overfitting.}
  \label{Training and testing MSE loss curves across}
\end{figure*}

To examine how DBLoss affects training dynamics and generalization capabilities, we use both MSE loss and DBLoss as objective functions and visualize the MSE on the training and testing datasets across all training epochs---see Figure~\ref{Training and testing MSE loss curves across}. We observe a consistent trend across all datasets. During training, models optimized with only MSE loss have lower errors per epoch compared to those optimized with DBLoss. However, on the test data, models trained with MSE loss exhibit higher errors per epoch than those trained with DBLoss. These observations indicate that while models trained with MSE loss have lower losses during the training phase, they generalize poorly on test data. In contrast, DBLoss enhances the model's generalization and prediction accuracy by encouraging the model to learn the trends and seasonal patterns in the dataset. Additionally, models trained with MSE loss show significant MSE fluctuations in the test loss on some datasets (e.g., Figure~\ref{Results on ETTm2 dataset} and Figure~\ref{Results on Solar dataset}), whereas DBLoss demonstrates greater stability.


\subsection{Hyperparameter Sensitivity}
Our method has two hyperparameters: the score weight \(\beta\) for weighted loss and the smoothing factor \(\alpha\) for EMA decomposition. To handle extreme cases, we manually replace $\alpha$ = 0 and $\alpha$ = 1 with approximate values close to 0 and 1, respectively. Specifically, a larger \(\beta\) increases the proportion of the seasonal component in the loss calculation, while a smaller \(\alpha\) results in heavier smoothing, making the trend smoother and the seasonal component more prominent. \begin{wrapfigure}{r}{0.6\columnwidth}
  \centering
  \raisebox{0pt}[\height][\depth]{\includegraphics[width=0.6\columnwidth]{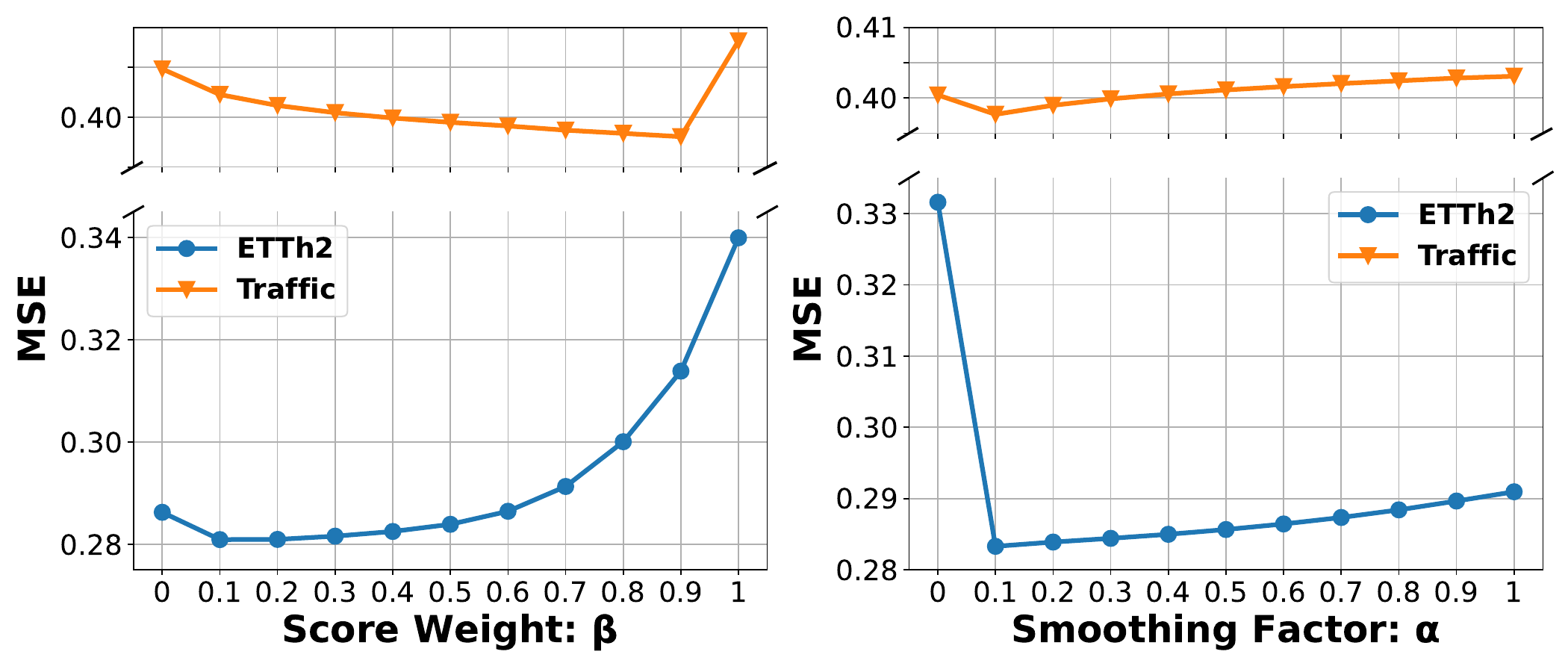}}
  \caption{The impact of the hyperparameter on ETTh2 and Traffic based DLinear (horizon 96).}
  \label{hyperparameter}
\end{wrapfigure}Conversely, a larger \(\alpha\) results in less smoothing, making the trend less smooth and the seasonal component less noticeable. From Figure~\ref{hyperparameter}, we have the following observations: 1) When \(\beta\) is too large (e.g., \(\beta = 1\)) or too small (e.g., \(\beta = 0\)), the model's performance is poor.
2) For datasets with pronounced seasonality, such as traffic, a larger score weight \(\beta\) (i.e., considering a higher proportion of the seasonal component in the loss calculation) yields better performance. A smaller smoothing factor \(\alpha\) (i.e., making the seasonal component more prominent) also improves performance.
3) For datasets with less pronounced seasonality, such as ETTh2, a moderate \(\beta\) value (e.g., 0.4 or 0.5) achieves better results, indicating that the proportions of the seasonal and trend components should be balanced. The variation in the smoothing factor \(\alpha\) has a minimal impact on performance.
4) However, we find that the optimal values of $\alpha$ and $\beta$ may vary across different algorithms. At present, there is no definitive method for selecting these parameters. We discuss this limitation in Appendix~\ref{app: limit} and leave it as an open problem for future research.

\subsection{Forecasting Visualization}
\begin{figure*}[!h]
    \centering
    \includegraphics[width=1\linewidth]{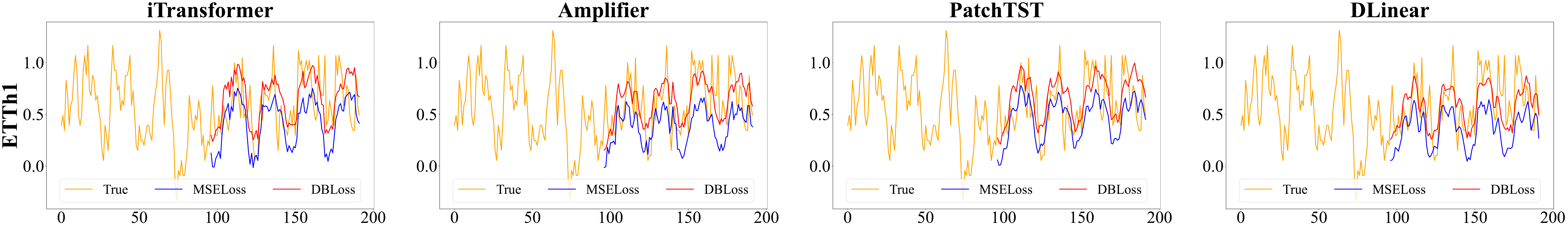}
    \caption{Forecasting visualization comparing DBLoss and MSE loss as objective functions.}
\label{Forecasting visualization comparing DBLoss and MSE loss as objective functions}
\end{figure*}

Figure~\ref{Forecasting visualization comparing DBLoss and MSE loss as objective functions} shows the visualization of forecasting results for samples from the ETTh1 dataset. We can observe that the predictions obtained with DBLoss are closer to the ground truth. This is mainly because DBLoss encourages the model to better learn the seasonal and trend patterns in the dataset. More visualization results are provided in Appendix~\ref{VISUALIZATION}.


\section{Conclusion}
In this study, we propose DBLoss to address the traditional MSE that sometimes fails to accurately capture the seasonality or trend within the forecasting horizon, even when decomposition modules are used in the forward propagation to model the trend and seasonality separately. Specifically, our method uses exponential moving averages to decompose the time series into seasonal and trend components within the forecasting horizon, and then calculates the loss for each of these components separately, followed by weighting them. By introducing DBLoss into the baseline model, we have achieved performance that surpasses the state-of-the-art on eight real-world datasets. Additionally, all datasets and code are available at \href{https://github.com/decisionintelligence/DBLoss}{https://github.com/decisionintelligence/DBLoss}.

\clearpage
\begin{ack}
This work was partially supported by the National Natural Science Foundation of China (No. 62472174), the Open Research Fund of Key Laboratory of Advanced Theory and Application in Statistics and Data Science–MOE, ECNU, and the Fundamental Research Funds for the Central Universities.
\end{ack}

\bibliography{reference}
\bibliographystyle{unsrtnat}

\newpage
\section*{NeurIPS Paper Checklist}

\begin{enumerate}

\item {\bf Claims}
    \item[] Question: Do the main claims made in the abstract and introduction accurately reflect the paper's contributions and scope?
    \item[] Answer: \answerYes{} 
    \item[] Justification: The main claims in the abstract and introduction accurately reflect our contributions.

\item {\bf Limitations}
    \item[] Question: Does the paper discuss the limitations of the work performed by the authors?
    \item[] Answer: \answerYes{} 
    \item[] Justification: We discuss the limitations of DBLoss in the Appendix~\ref{app: limit}.

\item {\bf Theory assumptions and proofs}
    \item[] Question: For each theoretical result, does the paper provide the full set of assumptions and a complete (and correct) proof?
    \item[] Answer: \answerYes{} 
    \item[] Justification: This paper includes the theoretical analysis.

    \item {\bf Experimental result reproducibility}
    \item[] Question: Does the paper fully disclose all the information needed to reproduce the main experimental results of the paper to the extent that it affects the main claims and/or conclusions of the paper (regardless of whether the code and data are provided or not)?
    \item[] Answer: \answerYes{} 
    \item[] Justification: We provide complete experimental details in Section~\ref{Setup}. Additionally, we have shared the full reproducible code and datasets in an anonymous repository (link provided under the abstract).

\item {\bf Open access to data and code}
    \item[] Question: Does the paper provide open access to the data and code, with sufficient instructions to faithfully reproduce the main experimental results, as described in supplemental material?
    \item[] Answer: \answerYes{} 
    \item[] Justification: We provide an anonymous link to the code (under the abstract) and describe how to reproduce the
    experimental results in the README file of the code.

\item {\bf Experimental setting/details}
    \item[] Question: Does the paper specify all the training and test details (e.g., data splits, hyperparameters, how they were chosen, type of optimizer, etc.) necessary to understand the results?
    \item[] Answer: \answerYes{} 
    \item[] Justification: We describe the complete experimental details in Section~\ref{Setup}.

\item {\bf Experiment statistical significance}
    \item[] Question: Does the paper report error bars suitably and correctly defined or other appropriate information about the statistical significance of the experiments?
    \item[] Answer: \answerYes{} 
    \item[] Justification: We report the standard deviations of the results for all the loss functions under
    different settings in Appendix~\ref{Comparison with Other Loss Functions Appendix}.

\item {\bf Experiments compute resources}
    \item[] Question: For each experiment, does the paper provide sufficient information on the computer resources (type of compute workers, memory, time of execution) needed to reproduce the experiments?
    \item[] Answer: \answerNA{} 
    \item[] Justification: The proposed DBLoss is generally applicable to arbitrary deep neural networks with negligible cost.
    
\item {\bf Code of ethics}
    \item[] Question: Does the research conducted in the paper conform, in every respect, with the NeurIPS Code of Ethics \url{https://neurips.cc/public/EthicsGuidelines}?
    \item[] Answer: \answerYes{} 
    \item[] Justification: 
    Our research aligns with the NeurIPS Code of Ethics.

\item {\bf Broader impacts}
    \item[] Question: Does the paper discuss both potential positive societal impacts and negative societal impacts of the work performed?
    \item[] Answer: \answerNA{} 
    \item[] Justification: 
    The paper focuses on advancing the field of machine learning. While our work may have various societal implications, we believe none are significant enough to warrant specific mention here.
    
\item {\bf Safeguards}
    \item[] Question: Does the paper describe safeguards that have been put in place for responsible release of data or models that have a high risk for misuse (e.g., pretrained language models, image generators, or scraped datasets)?
    \item[] Answer: \answerNA{} 
    \item[] Justification: 
    This paper poses no such risks.

\item {\bf Licenses for existing assets}
    \item[] Question: Are the creators or original owners of assets (e.g., code, data, models), used in the paper, properly credited and are the license and terms of use explicitly mentioned and properly respected?
    \item[] Answer: \answerYes{} 
    \item[] Justification: 
    The code and datasets used in the paper are publicly available and properly credited.

\item {\bf New assets}
    \item[] Question: Are new assets introduced in the paper well documented and is the documentation provided alongside the assets?
    \item[] Answer: \answerYes{} 
    \item[] Justification: 
    We will make the code publicly available upon acceptance of the paper and provide detailed documentation.

\item {\bf Crowdsourcing and research with human subjects}
    \item[] Question: For crowdsourcing experiments and research with human subjects, does the paper include the full text of instructions given to participants and screenshots, if applicable, as well as details about compensation (if any)? 
    \item[] Answer: \answerNA{} 
    \item[] Justification: 
    This work does not involve crowdsourcing nor research with human subjects.

\item {\bf Institutional review board (IRB) approvals or equivalent for research with human subjects}
    \item[] Question: Does the paper describe potential risks incurred by study participants, whether such risks were disclosed to the subjects, and whether Institutional Review Board (IRB) approvals (or an equivalent approval/review based on the requirements of your country or institution) were obtained?
    \item[] Answer: \answerNA{} 
    \item[] Justification: 
    This work does not involve crowdsourcing nor research with human subjects.

\item {\bf Declaration of LLM usage}
    \item[] Question: Does the paper describe the usage of LLMs if it is an important, original, or non-standard component of the core methods in this research? Note that if the LLM is used only for writing, editing, or formatting purposes and does not impact the core methodology, scientific rigorousness, or originality of the research, declaration is not required.
    \item[] Answer: \answerNA{} 
    \item[] Justification: 
    Our propsosed method does not include any component related to LLMs.

\end{enumerate}

\clearpage
\appendix
\section{Datasets}
\label{appendix dataset}

To conduct comprehensive and fair comparisons for different models, we conduct experiments on 
eight well-known forecasting benchmarks as the target datasets, including (I) \textbf{ETT} (Electricity Transformer Temperature, 4 subsets) data contains seven features of electricity transformer data collected from two separate counties between July 2016 and July 2018. These datasets vary in granularity, with ``h'' indicating hourly data and ``m'' indicating 15-minute intervals. The suffixes ``1'' and ``2'' refer to two different regions from which the datasets originated. (II) \textbf{Weather} data includes 21 meteorological factors recorded every 10 minutes in 2020 at the Max Planck Biogeochemistry Institute’s Weather Station. (III) \textbf{Electricity} data contains hourly electricity consumption data of 321 clients from 2012 to 2014. (IV) \textbf{Solar} data records the solar power production of 137 PV plants in 2006, which are sampled every 10 minutes. (V) \textbf{Traffic} data contains road occupancy rates measured by 862 sensors on freeways in the San Francisco Bay Area from 2015 to 2016, recorded hourly.

\renewcommand{\arraystretch}{1} %
\begin{table*}[!h]
\caption{Statistics of datasets.}
\label{Multivariate datasets}
\resizebox{1\columnwidth}{!}{
\begin{tabular}{@{}lllrrcl@{}}
\toprule
Dataset      & Domain      & Frequency & Lengths & Dim & Split  & Description\\ \midrule
ETTh1        & Electricity & 1 hour     & 14,400      & 7        & 6:2:2 & Power transformer 1, comprising seven indicators such as oil temperature and useful load\\
ETTh2        & Electricity & 1 hour    & 14,400      & 7        & 6:2:2 & Power transformer 2, comprising seven indicators such as oil temperature and useful load\\
ETTm1        & Electricity & 15 mins   & 57,600      & 7        & 6:2:2 & Power transformer 1, comprising seven indicators such as oil temperature and useful load\\
ETTm2        & Electricity & 15 mins   & 57,600      & 7        & 6:2:2 & Power transformer 2, comprising seven indicators such as oil temperature and useful load\\
Weather      & Environment & 10 mins   & 52,696      & 21       & 7:1:2 & Recorded
every for the whole year 2020, which contains 21 meteorological indicators\\
Electricity  & Electricity & 1 hour    & 26,304      & 321      & 7:1:2 & Electricity records the electricity consumption in kWh every 1 hour from 2012 to 2014\\
Solar        & Energy      & 10 mins   & 52,560      & 137      & 6:2:2 &Solar production records collected from 137 PV plants in Alabama \\
Traffic      & Traffic     & 1 hour    & 17,544      & 862      & 7:1:2 & Road occupancy rates measured by 862 sensors on San Francisco Bay area freeways\\
 \bottomrule
\end{tabular}}
\end{table*}


\section{Related Works}
Time series forecasting (TSF) predicts future observations based on historical observations. TSF methods are mainly categorized into four distinct approaches: (1) statistical learning-based methods, (2) machine learning-based methods, (3) deep learning-based methods, and (4) foundation methods. 

Early TSF methods primarily rely on statistical learning approaches such as ARIMA~\citep{box1970distribution}, ETS~\citep{hyndman2008forecasting}, and VAR~\citep{godahewa2021monash}. With advancements in machine learning, methods like XGBoost~\citep{chen2016xgboost}, Random Forests~\citep{breiman2001random}, and LightGBM~\citep{ke2017lightgbm} gain popularity for handling nonlinear patterns. However, these methods still require manual feature engineering and model design~\citep{AutoCTS++,lu2024robust,encoder,PAIR,lu2025does,li2025multi,zhou2025reagent}. Leveraging the representation learning of deep neural networks (DNNs)~\citep{qiu2025comprehensive,FineCIR,MEDIAN,yang2024wcdt, lu2023tf,wang2025unitmge}, many deep learning-based methods emerge. TimesNet~\citep{TimesNet} and SegRNN~\citep{lin2023segrnn} model time series as vector sequences, using CNNs or RNNs to capture temporal dependencies. Additionally, Transformer architectures, including DUET~\citep{DUET}, Informer~\citep{zhou2021informer}, DAG~\citep{DAG}, FEDformer~\citep{Fedformer}, Triformer~\citep{Triformer}, and PatchTST~\citep{PatchTST}, capture complex relationships between time points more accurately, significantly improving forecasting performance. MLP-based methods, including Hdmixer~\citep{huang2024hdmixer}, SparseTSF~\citep{lin2024sparsetsf}, CycleNet~\citep{lincyclenet}, APN~\citep{APN}, SRSNet~\citep{wu2025srsnet}, NLinear~\citep{DLinear}, and DLinear~\citep{DLinear}, adopt simpler architectures with fewer parameters but still achieve highly competitive forecasting accuracy. 

However, many of these methods struggle with generalization across domains due to their reliance on domain-specific data~\citep{FM4TS-Bench}. To address this, foundation methods are proposed, categorized into LLM-based methods and time series pre-trained methods. LLM-based methods~\citep{gpt4ts, time-llm, unitime, S2IP-LLM} leverage the strong representational capacity and sequential modeling capability of LLMs to capture complex patterns for time series modeling. Time series pre-trained methods~\citep{Timer, units, Moment, timesfm} focus on pre-training over multi-domain time series data, enabling the method to learn domain-agnostic features that are transferable across various applications. This strategy not only enhances performance on specific tasks but also provides greater flexibility when adapting to new datasets or scenarios.

\section{Theoretical Proofs}
\label{Theoretical Proofs}
In this section, we provide a theoretical analysis to explain why the proposed DBLoss is more effective than the conventional MSE loss in time series forecasting.

Motivated by the success of recent methods such as DLinear~\cite{DLinear}, DUET~\cite{DUET}, TimeMixer~\cite{TimeMixer}, and xPatch~\cite{xPatch}, 
which model time series by decomposing them into trend and seasonal components, achieving excellent performance, we assume that the trend and seasonal components are highly independent.. 

\subsection{Problem Formulation}

Let the original time series be denoted as $y_t$, which can be decomposed into a trend component $T_t$ and a seasonal component $S_t$:
\begin{equation}
y_t = T_t + S_t.
\end{equation}
Similarly, the model prediction $\hat{y}_t$ can be expressed as:
\begin{equation}
\hat{y}_t = \hat{T}_t + \hat{S}_t.
\end{equation}

\subsection{Analysis of MSE Loss}

Under this setting, the Mean Squared Error (MSE) can be expanded as:
\begin{align}
L_{\text{MSE}} 
&= \| y_t - \hat{y}_t \|_2^2 
= \| (T_t + S_t) - (\hat{T}_t + \hat{S}_t) \|_2^2 \nonumber \\
&= \| (T_t - \hat{T}_t) + (S_t - \hat{S}_t) \|_2^2 \nonumber \\
&= \| T_t - \hat{T}_t \|_2^2 + \| S_t - \hat{S}_t \|_2^2 
+ 2 \cdot (T_t - \hat{T}_t)(S_t - \hat{S}_t).
\end{align}

The key part from MSE lies in the cross term $2 \cdot (T_t - \hat{T}_t)(S_t - \hat{S}_t)$.  
Our assumption is that the trend and seasonal components are highly independent.However, this cross term introduces an interaction between them, potentially making it difficult for the model to optimize the two components independently, which can degrade the overall prediction performance. For instance, if the trend component is poorly predicted while the seasonal component is well captured, the interaction term can still yield a large negative value of $2 \cdot (T_t - \hat{T}_t)(S_t - \hat{S}_t)$, disproportionately affecting the total loss.

\subsection{Gradient Analysis of MSE}

We further analyze the MSE loss from the perspective of gradient propagation, and reveal MSE loss being unable to independently consider these two components during the optimization process. 
According to the chain rule:
\begin{align}
\nabla_{\boldsymbol{\Theta}} L_t 
&= 2 \cdot [(T_t - \hat{T}_t) + (S_t - \hat{S}_t)] 
\cdot \nabla_{\boldsymbol{\Theta}} (-\hat{T}_t - \hat{S}_t) \nonumber \\
&= -2 \cdot [(T_t - \hat{T}_t) + (S_t - \hat{S}_t)] 
\cdot (\nabla_{\boldsymbol{\Theta}} \hat{T}_t + \nabla_{\boldsymbol{\Theta}} \hat{S}_t).
\end{align}

Let the Jacobians of the trend and seasonal components be:
\begin{align}
\mathbf{J}_T := \nabla_{\boldsymbol{\Theta}} \hat{T}_t, \quad
\mathbf{J}_S := \nabla_{\boldsymbol{\Theta}} \hat{S}_t.
\end{align}

Then the gradient can be expressed as:
\begin{align}
\nabla_{\boldsymbol{\Theta}} L_t 
&= -2 \cdot \left[(T_t - \hat{T}_t)\mathbf{J}_T + (T_t - \hat{T}_t)\mathbf{J}_S 
+ (S_t - \hat{S}_t)\mathbf{J}_T + (S_t - \hat{S}_t)\mathbf{J}_S \right].
\end{align}

We decompose this gradient into two parts:

\paragraph{(1) Ideal Decoupled Term}
\begin{equation}
\mathbf{G}_{\text{ideal}} = -2 \cdot \left[(T_t - \hat{T}_t)\mathbf{J}_T + (S_t - \hat{S}_t)\mathbf{J}_S \right],
\end{equation}
which represents the desired independent optimization of the two components.

\paragraph{(2) Coupled Term}
\begin{equation}
\mathbf{G}_{\text{coupling}} = -2 \cdot \left[(T_t - \hat{T}_t)\mathbf{J}_S + (S_t - \hat{S}_t)\mathbf{J}_T \right].
\end{equation}

The coupled term $\mathbf{G}_{\text{coupling}}$ causes mutual interference:
the trend error influences the seasonal optimization and vice versa.
As long as $T_t \neq \hat{T}_t$ or $S_t \neq \hat{S}_t$ (i.e., the model has not converged),
$\mathbf{G}_{\text{coupling}} \neq \mathbf{0}$, meaning that the optimization of one component will inevitably affect the other.

\subsection{Analysis of DBLoss}
\begin{equation}
L_{\text{DB}} = \beta \cdot \|\hat{S}_t - S_t \|_2^2 + (1 - \beta) \cdot \| \hat{T}_t  - T_t\|_1,
\end{equation}
where $\beta$ is hyperparameters controlling the relative weights of the trend and seasonal components.

Unlike MSE, DBLoss explicitly separates the optimization of the two components, 
thus removing the coupling term $\mathbf{G}_{\text{coupling}}$ from the gradient computation.  
Furthermore, by adjusting the coefficients or using different distance norms, one can precisely control the loss scale for each component, enabling targeted learning and better modeling of both parts.

\section{Visualization}
\label{VISUALIZATION}
\begin{figure*}[!h]
  \centering
  \subfloat[Forecasting visualization on ETTh1 dataset]
  {\includegraphics[width=1\textwidth]{Figures/cherry0.pdf}\label{Results on cherry0 dataset}}
  
  \subfloat[Forecasting visualization on ETTm2 dataset]
  {\includegraphics[width=1\textwidth]{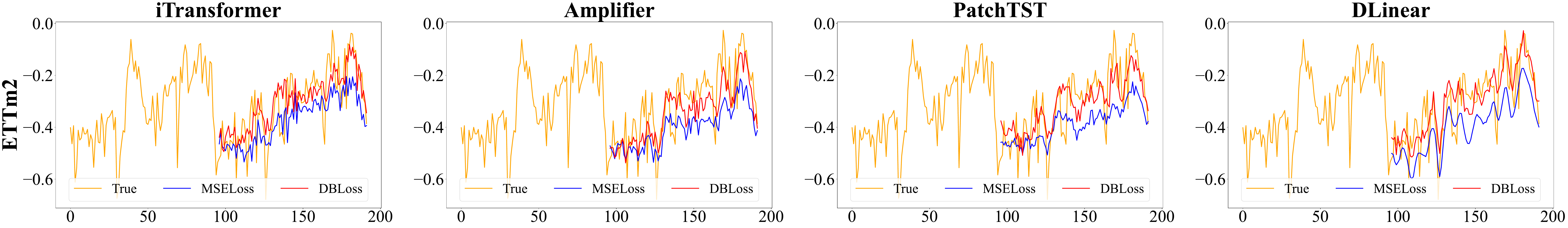}\label{Results on cherry1 dataset}}

  \subfloat[Forecasting visualization on Weather dataset]
  {\includegraphics[width=1\textwidth]{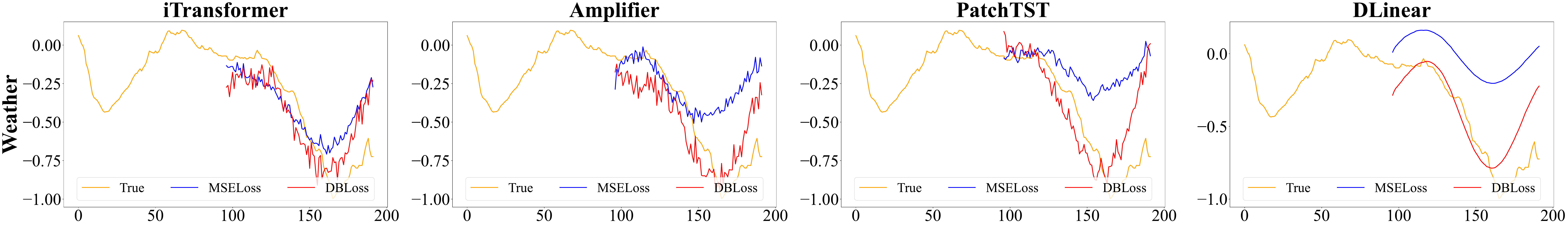}\label{Results on cherry3 dataset}}
  
  \subfloat[Forecasting visualization on Solar dataset]
  {\includegraphics[width=1\textwidth]{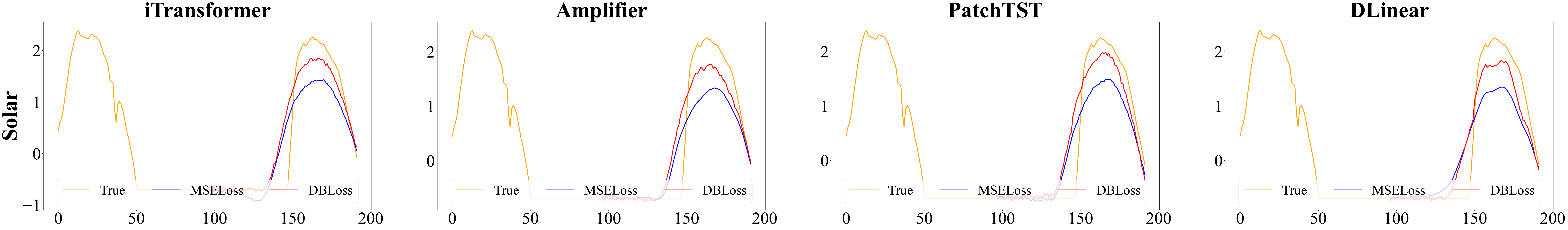}\label{Results on cherry2 dataset}}
  \caption{Forecasting visualization comparing DBLoss and MSE loss as objective functions.}
  \label{Forecasting visualization}
\end{figure*}

\clearpage

\section{Comparison with Other Loss Functions}
\label{Comparison with Other Loss Functions Appendix}

\begin{table}[h]
\centering
  \caption{Comparison between the proposed DBLoss and other loss functions. The model is DLinear and we report the result of ETTh2. The best results are highlighted in \textbf{bold}, and the second-best results are highlighted in \underline{underline}. The standard deviation of methods calculated through 5 random seeds are also reported.}
  \resizebox{0.7\columnwidth}{!}{%
\begin{tabular}{cc|ccccc}
    \toprule
 \multicolumn{2}{c|}{Dataset} & \multicolumn{5}{c}{ETTh2} \\    \midrule
 \multicolumn{2}{c|}{Forecast   horizon} & 96 & 192 & 336 & 720 & Avg\\    \midrule
 \multirow{2}[2]{*}{Ori}  & MSE & $0.300\scriptstyle\pm 0.002$ & $0.387\scriptstyle\pm 0.001$ & $0.490\scriptstyle\pm 0.001$ & $0.704\scriptstyle\pm 0.004$ & $0.470\scriptstyle\pm 0.001$ \\
 & MAE & $0.364\scriptstyle\pm 0.002$ & $0.423\scriptstyle\pm 0.002$ & $0.487\scriptstyle\pm 0.002$ & $0.597\scriptstyle\pm 0.001$ & $0.468\scriptstyle\pm 0.002$ \\ \midrule
TILDE-Q & MSE & $0.287\scriptstyle\pm 0.002$ & $0.362\scriptstyle\pm 0.001$ & $0.425\scriptstyle\pm 0.002$ & $0.599\scriptstyle\pm 0.001$ & $0.418\scriptstyle\pm 0.002$ \\
(\citeyear{TILDE-Q}) & MAE & $0.345\scriptstyle\pm 0.004$ & $0.395\scriptstyle\pm 0.001$ & $0.445\scriptstyle\pm 0.002$ & $0.551\scriptstyle\pm 0.002$ & $0.434\scriptstyle\pm 0.001$ \\ \midrule
FreDF & MSE & $\underline{0.284}\scriptstyle\pm 0.001$ & $0.362\scriptstyle\pm 0.001$ & $0.420\scriptstyle\pm 0.002$ & $\underline{0.587}\scriptstyle\pm 0.002$ & $\underline{0.413}\scriptstyle\pm 0.001$ \\
(\citeyear{FreDF}) & MAE & $\textbf{0.342}\scriptstyle\pm 0.003$ & $0.396\scriptstyle\pm 0.002$ & $0.445\scriptstyle\pm 0.004$ & $\underline{0.546}\scriptstyle\pm 0.003$ & $0.432\scriptstyle\pm 0.003$\\ \midrule
PSLoss & MSE & $\textbf{0.283}\scriptstyle\pm 0.002$ & $\underline{0.358}\scriptstyle\pm 0.002$ & $\underline{0.411}\scriptstyle\pm 0.002$ & $0.614\scriptstyle\pm 0.003$ & $0.417\scriptstyle\pm 0.002$ \\
(\citeyear{PSLoss}) & MAE & $0.343\scriptstyle\pm 0.003$ & $\underline{0.393}\scriptstyle\pm 0.004$ & $\underline{0.434}\scriptstyle\pm 0.002$ & $0.549\scriptstyle\pm 0.003$ & $\underline{0.430}\scriptstyle\pm 0.002$ \\ \midrule
DBLoss & MSE & $\underline{0.284}\scriptstyle\pm 0.002$ & $\textbf{0.357}\scriptstyle\pm 0.003$ & $\textbf{0.407}\scriptstyle\pm 0.002$ & $\textbf{0.586}\scriptstyle\pm 0.001$ & $\textbf{0.409}\scriptstyle\pm 0.001$ \\
(Ours) & MAE & $\textbf{0.342}\scriptstyle\pm 0.001$ & $\textbf{0.390}\scriptstyle\pm 0.001$ & $\textbf{0.430}\scriptstyle\pm 0.001$ & $\textbf{0.533}\scriptstyle\pm 0.004$ & $\textbf{0.424}\scriptstyle\pm 0.002$\\
 \bottomrule
\end{tabular}}
\end{table}

\begin{table}[h]
\centering
  \caption{Comparison between the proposed DBLoss and other loss functions. The model is DLinear and we report the result of ETTm1. The best results are highlighted in \textbf{bold}, and the second-best results are highlighted in \underline{underline}. The standard deviation of methods calculated through 5 random seeds are also reported.}
  \resizebox{0.7\columnwidth}{!}{%
\begin{tabular}{cc|ccccc}
    \toprule
 \multicolumn{2}{c|}{Dataset} & \multicolumn{5}{c}{ETTm1} \\    \midrule
 \multicolumn{2}{c|}{Forecast   horizon} & 96 & 192 & 336 & 720 & Avg\\    \midrule
 \multirow{2}[2]{*}{Ori}  & MSE & $0.300\scriptstyle\pm 0.003$ & $0.336\scriptstyle\pm 0.002$ & $0.367\scriptstyle\pm 0.002$ & $0.419\scriptstyle\pm 0.003$ & $0.356\scriptstyle\pm 0.005$ \\
 & MAE & $0.345\scriptstyle\pm 0.002$ & $0.366\scriptstyle\pm 0.002$ & $0.386\scriptstyle\pm 0.003$ & $0.416\scriptstyle\pm 0.003$ & $0.378\scriptstyle\pm 0.003$ \\\midrule
TILDE-Q & MSE & $0.302\scriptstyle\pm 0.002$ & $0.336\scriptstyle\pm 0.002$ & $0.371\scriptstyle\pm 0.002$ & $0.425\scriptstyle\pm 0.003$ & $0.359\scriptstyle\pm 0.003$ \\
(\citeyear{TILDE-Q}) & MAE & $0.342\scriptstyle\pm 0.005$ & $0.362\scriptstyle\pm 0.003$ & $0.386\scriptstyle\pm 0.002$ & $0.417\scriptstyle\pm 0.002$ & $0.377\scriptstyle\pm 0.002$ \\\midrule
FreDF & MSE & $0.302\scriptstyle\pm 0.003$ & $0.333\scriptstyle\pm 0.004$ & $0.363\scriptstyle\pm 0.001$ & $\textbf{0.415}\scriptstyle\pm 0.001$ & $0.353\scriptstyle\pm 0.003$ \\
(\citeyear{FreDF}) & MAE & $0.344\scriptstyle\pm 0.001$ & $0.363\scriptstyle\pm 0.001$ & $0.381\scriptstyle\pm 0.003$ & $\underline{0.411}\scriptstyle\pm 0.002$ & $0.375\scriptstyle\pm 0.002$ \\\midrule
PSLoss & MSE & $\underline{0.296}\scriptstyle\pm 0.001$ & $\underline{0.332}\scriptstyle\pm 0.003$ & $\textbf{0.361}\scriptstyle\pm 0.001$ & $0.416\scriptstyle\pm 0.004$ & $\textbf{0.351}\scriptstyle\pm 0.001$ \\
(\citeyear{PSLoss}) & MAE & $\underline{0.339}\scriptstyle\pm 0.002$ & $\underline{0.361}\scriptstyle\pm 0.001$ & $\underline{0.380}\scriptstyle\pm 0.002$ & $0.413\scriptstyle\pm 0.001$ & $\underline{0.373}\scriptstyle\pm 0.001$ \\\midrule
DBLoss & MSE & $\textbf{0.295}\scriptstyle\pm 0.001$ & $\textbf{0.331}\scriptstyle\pm 0.002$ & $\textbf{0.361}\scriptstyle\pm 0.002$ & $\textbf{0.415}\scriptstyle\pm 0.001$ & $\textbf{0.351}\scriptstyle\pm 0.002$ \\
(Ours) & MAE & $\textbf{0.337}\scriptstyle\pm 0.001$ & $\textbf{0.358}\scriptstyle\pm 0.001$ & $\textbf{0.378}\scriptstyle\pm 0.001$ & $\textbf{0.409}\scriptstyle\pm 0.001$ & $\textbf{0.370}\scriptstyle\pm 0.002$\\
 \bottomrule
\end{tabular}}
\end{table}

\begin{table}[h]
\centering
  \caption{Comparison between the proposed DBLoss and other loss functions. The model is DLinear and we report the result of Traffic. The best results are highlighted in \textbf{bold}, and the second-best results are highlighted in \underline{underline}. The standard deviation of methods calculated through 5 random seeds are also reported.}
  \resizebox{0.7\columnwidth}{!}{%
\begin{tabular}{cc|ccccc}
    \toprule
 \multicolumn{2}{c|}{Dataset} & \multicolumn{5}{c}{Traffic} \\    \midrule
 \multicolumn{2}{c|}{Forecast   horizon} & 96 & 192 & 336 & 720 & Avg\\    \midrule
\multirow{2}[2]{*}{Ori}  & MSE & $\textbf{0.395}\scriptstyle\pm 0.001$ & $\textbf{0.407}\scriptstyle\pm 0.001$ & $0.417\scriptstyle\pm 0.001$ & $0.454\scriptstyle\pm 0.003$ & $\underline{0.418}\scriptstyle\pm 0.002$ \\
 & MAE & $0.275\scriptstyle\pm 0.002$ & $0.280\scriptstyle\pm 0.001$ & $0.286\scriptstyle\pm 0.001$ & $0.308\scriptstyle\pm 0.001$ & $0.287\scriptstyle\pm 0.002$ \\\midrule
TILDE-Q & MSE & $0.416\scriptstyle\pm 0.003$ & $0.422\scriptstyle\pm 0.004$ & $0.423\scriptstyle\pm 0.001$ & $0.461\scriptstyle\pm 0.002$ & $0.431\scriptstyle\pm 0.003$ \\
(\citeyear{TILDE-Q}) & MAE & $0.294\scriptstyle\pm 0.002$ & $0.296\scriptstyle\pm 0.001$ & $0.293\scriptstyle\pm 0.002$ & $0.316\scriptstyle\pm 0.003$ & $0.300\scriptstyle\pm 0.002$ \\\midrule
FreDF & MSE & $0.398\scriptstyle\pm 0.001$ & $0.408\scriptstyle\pm 0.004$ & $\underline{0.416}\scriptstyle\pm 0.002$ & $\underline{0.452}\scriptstyle\pm 0.001$ & $0.419\scriptstyle\pm 0.001$ \\
(\citeyear{FreDF}) & MAE & $\textbf{0.270}\scriptstyle\pm 0.001$ & $0.275\scriptstyle\pm 0.003$ & $0.280\scriptstyle\pm 0.002$ & $0.302\scriptstyle\pm 0.002$ & $0.282\scriptstyle\pm 0.001$ \\\midrule
PSLoss & MSE & $0.398\scriptstyle\pm 0.001$ & $0.408\scriptstyle\pm 0.001$ & $\underline{0.416}\scriptstyle\pm 0.003$ & $\underline{0.452}\scriptstyle\pm 0.005$ & $0.419\scriptstyle\pm 0.002$ \\
(\citeyear{PSLoss}) & MAE & $\textbf{0.270}\scriptstyle\pm 0.001$ & $\textbf{0.274}\scriptstyle\pm 0.001$ & $\textbf{0.279}\scriptstyle\pm 0.001$ & $\underline{0.299}\scriptstyle\pm 0.003$ & $\underline{0.281}\scriptstyle\pm 0.002$ \\\midrule
DBLoss & MSE & $\underline{0.396}\scriptstyle\pm 0.001$ & $\textbf{0.407}\scriptstyle\pm 0.001$ & $\textbf{0.415}\scriptstyle\pm 0.001$ & $\textbf{0.449}\scriptstyle\pm 0.005$ & $\textbf{0.417}\scriptstyle\pm 0.002$ \\
(Ours) & MAE & $\textbf{0.270}\scriptstyle\pm 0.001$ & $\textbf{0.274}\scriptstyle\pm 0.001$ & $\textbf{0.279}\scriptstyle\pm 0.001$ & $\textbf{0.298}\scriptstyle\pm 0.003$ & $\textbf{0.280}\scriptstyle\pm 0.002$\\
 \bottomrule
\end{tabular}}
\end{table}

\clearpage

\section{Efficiency}
Table~\ref{training time} presents the epoch-level training times (in seconds) of PatchTST when using DBLoss and MSE across different datasets. The results show the average values for the four forecasting horizons of each dataset, with the same parameters, where only MSE is replaced by DBLoss. Based on the experimental results in the table, we can observe that DBLoss does lead to an increase in training time compared to MSE, but this increase is not significant. As the dataset size grows, the time difference becomes even more negligible.
\begin{table}[h]
  \caption{Comparison of average epoch-level training times (in seconds) between DBLoss and MSE using PatchTST across four forecasting horizons on different datasets.}
  \label{training time}
  \resizebox{1\columnwidth}{!}{%
\begin{tabular}{c|cccccccc}
\toprule
\textbf{Train Time} & ETTh1 & ETTh2 & ETTm1 & ETTm2 & {Solar} & {Weather} & {Electricity} & {Traffic} \\\midrule
MSE & 2.36 & 2.37 & 14.45 & 14.39 & {183.11} & {36.07} & {258.47} & {1035.77} \\
DBLoss & 3.11 & 3.37 & 15.93 & 15.73 & {186.31} & {37.23} & {260.85} & {1039.67}\\
\bottomrule
\end{tabular}}
\end{table}


\section{Full Results on Time Series Foundation Models}
\label{Full Results on Time Series Foundation Models}
Table~\ref{Foundation models} presents the results of foundation models under the 5\% few-shot setting. It reports both MSE and MAE across different forecasting horizons $F \in {96, 192, 336, 720}$. The baseline parameters are kept consistent with those used in TSFM-Bench~\citep{li2025TSFM-Bench}. The best results are highlighted in \textbf{bold}.

\begin{table*}[t]
\caption{Foundation models results in the 5\% few-shot setting. The table reports MSE and MAE for different forecasting lengths $F \in \{96, 192, 336, 720\}$. The parameters for the baselines are kept consistent with those of TSFM-Bench~\citep{li2025TSFM-Bench}. The better results are highlighted in \textbf{bold}.}
\label{Foundation models}
\resizebox{\columnwidth}{!}{
\begin{tabular}{c|c|cccc|cccc|cccc|cccc}
\toprule
\multicolumn{2}{c|}{Model} & \multicolumn{4}{c|}{GPT4TS} & \multicolumn{4}{c|}{CALF} & \multicolumn{4}{c|}{TTM} & \multicolumn{4}{c}{UniTS} \\\midrule
\multicolumn{2}{c|}{Loss} & \multicolumn{2}{c}{Ori} & \multicolumn{2}{c|}{DBLoss} & \multicolumn{2}{c}{Ori} & \multicolumn{2}{c|}{DBLoss} & \multicolumn{2}{c}{Ori} & \multicolumn{2}{c|}{DBLoss} & \multicolumn{2}{c}{Ori} & \multicolumn{2}{c}{DBLoss} \\     \midrule
\multicolumn{2}{c|}{Metric} & MSE & MAE & MSE & MAE & MSE & MAE & MSE & MAE & MSE & MAE & MSE & MAE & MSE & MAE & MSE & MAE \\ \midrule
  \multirow[c]{5}{*}{\rotatebox{90}{ETTh1}}& {96} & \textbf{0.438} & \textbf{0.445} & 0.439 & 0.446 & 0.405 & 0.426 & \textbf{0.401} & \textbf{0.422} & 0.363 & 0.392 & \textbf{0.361} & \textbf{0.390} & \textbf{0.381} & 0.394 & \textbf{0.381} & \textbf{0.393} \\
 & {192} & 0.460 & 0.458 & \textbf{0.455} & \textbf{0.456} & 0.428 & 0.442 & \textbf{0.423} & \textbf{0.436} & 0.391 & 0.409 & \textbf{0.387} & \textbf{0.405} & 0.421 & 0.428 & \textbf{0.405} & \textbf{0.424} \\
 & {336} & 0.462 & 0.467 & \textbf{0.449} & \textbf{0.460} & 0.443 & 0.454 & \textbf{0.437} & \textbf{0.447} & 0.411 & 0.429 & \textbf{0.404} & \textbf{0.422} & 0.443 & 0.437 & \textbf{0.429} & \textbf{0.429} \\
 & {720} & 0.509 & 0.511 & \textbf{0.470} & \textbf{0.486} & 0.495 & 0.494 & \textbf{0.472} & \textbf{0.479} & 0.453 & 0.471 & \textbf{0.429} & \textbf{0.448} & 0.498 & 0.475 & \textbf{0.486} & \textbf{0.463} \\\cmidrule{2-18} 
 & {Avg} & 0.467 & 0.470 & \textbf{0.453} & \textbf{0.462} & 0.443 & 0.454 & \textbf{0.433} & \textbf{0.446} & 0.405 & 0.425 & \textbf{0.395} & \textbf{0.417} & 0.436 & 0.434 & \textbf{0.425} & \textbf{0.427} \\\midrule
 \multirow[c]{5}{*}{\rotatebox{90}{ETTh2}} & {96} & 0.329 & 0.380 & \textbf{0.323} & \textbf{0.371} & \textbf{0.302} & \textbf{0.362} & 0.308 & 0.363 & 0.271 & 0.329 & \textbf{0.270} & \textbf{0.328} & 0.305 & 0.353 & \textbf{0.299} & \textbf{0.346} \\
 & {192} & 0.368 & 0.406 & \textbf{0.364} & \textbf{0.399} & 0.385 & 0.400 & \textbf{0.383} & \textbf{0.397} & 0.339 & \textbf{0.373} & \textbf{0.329} & 0.374 & 0.369 & 0.403 & \textbf{0.357} & \textbf{0.390} \\
 & {336} & 0.378 & 0.421 & \textbf{0.374} & \textbf{0.412} & 0.387 & 0.418 & \textbf{0.375} & \textbf{0.412} & 0.372 & 0.401 & \textbf{0.346} & \textbf{0.392} & 0.388 & 0.412 & \textbf{0.361} & \textbf{0.399} \\
 & {720} & 0.418 & 0.450 & \textbf{0.412} & \textbf{0.442} & 0.416 & 0.449 & \textbf{0.407} & \textbf{0.445} & 0.385 & 0.428 & \textbf{0.382} & \textbf{0.419} & 0.425 & 0.451 & \textbf{0.410} & \textbf{0.439} \\\cmidrule{2-18} 
 & {Avg} & 0.373 & 0.414 & \textbf{0.368} & \textbf{0.406} & 0.373 & 0.407 & \textbf{0.368} & \textbf{0.404} & 0.342 & 0.383 & \textbf{0.332} & \textbf{0.378} & 0.372 & 0.405 & \textbf{0.357} & \textbf{0.393} \\\midrule
 \multirow[c]{5}{*}{\rotatebox{90}{ETTm1}} & {96} & 0.343 & 0.379 & \textbf{0.330} & \textbf{0.368} & 0.317 & 0.366 & \textbf{0.299} & \textbf{0.347} & 0.299 & 0.343 & \textbf{0.294} & \textbf{0.337} & 0.313 & 0.363 & \textbf{0.300} & \textbf{0.349} \\
 & {192} & 0.375 & 0.398 & \textbf{0.361} & \textbf{0.387} & 0.346 & 0.380 & \textbf{0.337} & \textbf{0.369} & 0.341 & 0.367 & \textbf{0.340} & \textbf{0.365} & 0.357 & 0.390 & \textbf{0.338} & \textbf{0.373} \\
 & {336} & 0.394 & 0.406 & \textbf{0.384} & \textbf{0.398} & 0.385 & 0.405 & \textbf{0.371} & \textbf{0.391} & 0.365 & 0.381 & \textbf{0.363} & \textbf{0.379} & 0.381 & 0.405 & \textbf{0.370} & \textbf{0.392} \\
 & {720} & 0.440 & 0.434 & \textbf{0.432} & \textbf{0.425} & 0.439 & 0.433 & \textbf{0.427} & \textbf{0.420} & 0.420 & 0.412 & \textbf{0.419} & \textbf{0.409} & 0.457 & 0.448 & \textbf{0.438} & \textbf{0.428} \\\cmidrule{2-18} 
& {Avg} & 0.388 & 0.404 & \textbf{0.377} & \textbf{0.394} & 0.372 & 0.396 & \textbf{0.358} & \textbf{0.382} & 0.356 & 0.376 & \textbf{0.354} & \textbf{0.372} & 0.377 & 0.402 & \textbf{0.362} & \textbf{0.386} \\\midrule
 \multirow[c]{5}{*}{\rotatebox{90}{ETTm2}} & {96} & 0.190 & 0.279 & \textbf{0.181} & \textbf{0.266} & 0.180 & 0.272 & \textbf{0.170} & \textbf{0.258} & 0.164 & 0.250 & \textbf{0.162} & \textbf{0.244} & 0.188 & 0.278 & \textbf{0.173} & \textbf{0.255} \\
 & {192} & 0.241 & 0.312 & \textbf{0.231} & \textbf{0.300} & 0.237 & 0.310 & \textbf{0.228} & \textbf{0.295} & \textbf{0.222} & 0.290 & 0.224 & \textbf{0.287} & 0.255 & 0.317 & \textbf{0.247} & \textbf{0.302} \\
 & {336} & 0.296 & 0.349 & \textbf{0.281} & \textbf{0.330} & 0.295 & 0.348 & \textbf{0.277} & \textbf{0.328} & 0.282 & 0.330 & \textbf{0.278} & \textbf{0.323} & 0.321 & 0.366 & \textbf{0.285} & \textbf{0.331} \\
 & {720} & 0.385 & 0.401 & \textbf{0.371} & \textbf{0.384} & 0.372 & 0.397 & \textbf{0.363} & \textbf{0.382} & 0.364 & 0.381 & \textbf{0.363} & \textbf{0.376} & 0.404 & 0.415 & \textbf{0.374} & \textbf{0.392} \\\cmidrule{2-18} 
 & {Avg} & 0.278 & 0.335 & \textbf{0.266} & \textbf{0.320} & 0.271 & 0.332 & \textbf{0.259} & \textbf{0.316} & 0.258 & 0.313 & \textbf{0.257} & \textbf{0.308} & 0.292 & 0.344 & \textbf{0.270} & \textbf{0.320} \\\midrule
 \multirow[c]{5}{*}{\rotatebox{90}{Solar}} & {96} & 0.253 & 0.326 & \textbf{0.243} & \textbf{0.267} & \textbf{0.203} & \textbf{0.274} & 0.238 & 0.295 & \textbf{0.201} & 0.254 & \textbf{0.201} & \textbf{0.244} & \textbf{0.186} & 0.244 & \textbf{0.186} & \textbf{0.226} \\
 & {192} & 0.266 & 0.336 & \textbf{0.252} & \textbf{0.281} & \textbf{0.224} & \textbf{0.290} & 0.243 & 0.294 & \textbf{0.225} & 0.270 & 0.235 & \textbf{0.267} & \textbf{0.198} & 0.255 & 0.206 & \textbf{0.241} \\
 & {336} & 0.262 & 0.341 & \textbf{0.260} & \textbf{0.284} & \textbf{0.243} & 0.308 & 0.252 & \textbf{0.303} & \textbf{0.222} & 0.274 & 0.231 & \textbf{0.271} & \textbf{0.208} & 0.259 & 0.222 & \textbf{0.254} \\
 & {720} & 0.265 & 0.335 & \textbf{0.261} & \textbf{0.285} & \textbf{0.247} & 0.314 & 0.251 & \textbf{0.306} & \textbf{0.226} & \textbf{0.277} & 0.229 & 0.283 & \textbf{0.231} & 0.284 & 0.243 & \textbf{0.265} \\\cmidrule{2-18} 
& {Avg} & 0.262 & 0.335 & \textbf{0.254} & \textbf{0.279} & \textbf{0.229} & \textbf{0.297} & 0.246 & 0.300 & \textbf{0.219} & 0.269 & 0.224 & \textbf{0.266} & \textbf{0.206} & 0.261 & 0.214 & \textbf{0.246} \\\midrule
 \multirow[c]{5}{*}{\rotatebox{90}{Weather}} & {96} & 0.187 & 0.244 & \textbf{0.181} & \textbf{0.234} & 0.163 & 0.217 & \textbf{0.162} & \textbf{0.210} & \textbf{0.147} & 0.195 & 0.148 & \textbf{0.191} & \textbf{0.154} & 0.206 & 0.156 & \textbf{0.198} \\
 & {192} & 0.225 & 0.274 & \textbf{0.220} & \textbf{0.265} & 0.206 & 0.253 & \textbf{0.205} & \textbf{0.250} & \textbf{0.194} & 0.238 & \textbf{0.194} & \textbf{0.233} & \textbf{0.199} & 0.248 & \textbf{0.199} & \textbf{0.237} \\
 & {336} & 0.268 & 0.304 & \textbf{0.265} & \textbf{0.297} & 0.260 & 0.297 & \textbf{0.257} & \textbf{0.292} & 0.244 & 0.277 & \textbf{0.243} & \textbf{0.272} & \textbf{0.248} & 0.285 & \textbf{0.248} & \textbf{0.275} \\
 & {720} & 0.330 & 0.348 & \textbf{0.327} & \textbf{0.340} & \textbf{0.322} & 0.339 & \textbf{0.322} & \textbf{0.338} & \textbf{0.314} & 0.329 & 0.316 & \textbf{0.327} & \textbf{0.320} & 0.337 & \textbf{0.320} & \textbf{0.329} \\\cmidrule{2-18} 
 & {Avg} & 0.253 & 0.293 & \textbf{0.248} & \textbf{0.284} & 0.238 & 0.277 & \textbf{0.236} & \textbf{0.272} & \textbf{0.225} & 0.260 & \textbf{0.225} & \textbf{0.256} & \textbf{0.230} & 0.269 & 0.231 & \textbf{0.260} \\\midrule
 \multirow[c]{5}{*}{\rotatebox{90}{Electricity}} & {96} & \textbf{0.178} & 0.294 & 0.179 & \textbf{0.286} & 0.141 & 0.240 & \textbf{0.140} & \textbf{0.236} & \textbf{0.146} & 0.246 & \textbf{0.146} & \textbf{0.245} & \textbf{0.150} & 0.249 & 0.152 & \textbf{0.248} \\
 & {192} & \textbf{0.192} & 0.306 & \textbf{0.192} & \textbf{0.300} & 0.156 & 0.254 & \textbf{0.154} & \textbf{0.250} & \textbf{0.165} & 0.264 & \textbf{0.165} & \textbf{0.262} & \textbf{0.167} & 0.264 & 0.168 & \textbf{0.263} \\
 & {336} & \textbf{0.208} & 0.318 & \textbf{0.208} & \textbf{0.310} & \textbf{0.174} & 0.271 & \textbf{0.174} & \textbf{0.268} & 0.181 & 0.281 & \textbf{0.180} & \textbf{0.278} & \textbf{0.181} & 0.277 & 0.182 & \textbf{0.276} \\
 & {720} & \textbf{0.248} & 0.348 & \textbf{0.248} & \textbf{0.339} & 0.216 & 0.306 & \textbf{0.214} & \textbf{0.302} & \textbf{0.223} & 0.315 & \textbf{0.223} & \textbf{0.313} & \textbf{0.220} & \textbf{0.309} & 0.224 & 0.310 \\\cmidrule{2-18} 
 & {Avg} & \textbf{0.207} & 0.317 & \textbf{0.207} & \textbf{0.309} & 0.172 & 0.268 & \textbf{0.171} & \textbf{0.264} & 0.179 & 0.277 & \textbf{0.178} & \textbf{0.274} & \textbf{0.180} & 0.275 & 0.181 & \textbf{0.274} \\\midrule
\multirow[c]{5}{*}{\rotatebox{90}{Traffic}} & {96} & 0.411 & 0.300 & \textbf{0.408} & \textbf{0.286} & 0.406 & 0.298 & \textbf{0.405} & \textbf{0.290} & 0.448 & 0.324 & \textbf{0.445} & \textbf{0.322} & 0.401 & 0.278 & \textbf{0.400} & \textbf{0.272} \\
 & {192} & 0.422 & 0.304 & \textbf{0.416} & \textbf{0.289} & 0.423 & 0.309 & \textbf{0.420} & \textbf{0.301} & 0.466 & 0.330 & \textbf{0.463} & \textbf{0.329} & 0.414 & 0.284 & \textbf{0.412} & \textbf{0.278} \\
 & {336} & 0.432 & 0.308 & \textbf{0.426} & \textbf{0.293} & 0.436 & 0.317 & \textbf{0.432} & \textbf{0.310} & 0.491 & 0.345 & \textbf{0.487} & \textbf{0.343} & 0.421 & 0.290 & \textbf{0.417} & \textbf{0.280} \\
 & {720} & 0.468 & 0.325 & \textbf{0.463} & \textbf{0.311} & 0.477 & 0.340 & \textbf{0.476} & \textbf{0.335} & 0.533 & 0.365 & \textbf{0.527} & \textbf{0.361} & 0.452 & 0.305 & \textbf{0.451} & \textbf{0.298} \\\cmidrule{2-18} 
 & {Avg} & 0.433 & 0.309 & \textbf{0.428} & \textbf{0.295} & 0.435 & 0.316 & \textbf{0.433} & \textbf{0.309} & 0.484 & 0.341 & \textbf{0.481} & \textbf{0.339} & 0.422 & 0.289 & \textbf{0.420} & \textbf{0.282}\\\bottomrule
\end{tabular}}
\end{table*}

\section{Limitations}
\label{app: limit}
\textbf{Potential limitations} The DBLoss demonstrates its efficacy in TSF scenarios. However, there are several potential limitations of DBLoss
that warrant discussion here:
\begin{itemize}
    \item \textbf{Lack of Automated Hyperparameter Tuning Strategy:} The proposed method involves two critical hyperparameters: the score weight \(\beta\) for the weighted loss and the smoothing factor \(\alpha\) for Exponential Moving Average (EMA) decomposition. Specifically, a larger \(\beta\) increases the proportion of the seasonal component in the loss calculation, while a smaller \(\alpha\) results in stronger smoothing, making the trend smoother and the seasonal component more prominent. However, the current research has not yet proposed an automated strategy to optimize the selection of these two parameters. The absence of a systematic tuning approach may still limit the model's performance improvement and generalization capability. Therefore, developing an efficient automated hyperparameter tuning mechanism to adaptively determine the optimal parameter combination is an important direction for future research.
    
\end{itemize}

\end{document}